\documentclass[11pt,a4paper]{article}
\usepackage{times,latexsym}
\usepackage{url}
\usepackage[T1]{fontenc}
\usepackage{amsmath}
\usepackage{amsfonts}
\usepackage{tabularx}
\usepackage{subcaption}
\usepackage[most]{tcolorbox}
\usepackage[rightcaption]{sidecap}
\tcbuselibrary{listingsutf8}
\tcbset{
  colback=gray!5!white, colframe=black!75!black,
  fonttitle=\bfseries, boxed title style={colback=black!15!white}
}

\usepackage[acceptedWithA]{tacl2021v1}

\usepackage{xspace,mfirstuc,tabulary}

\usepackage{graphicx}
\usepackage{booktabs}
\usepackage{caption}

\newif\iftaclinstructions
\taclinstructionsfalse %
\iftaclinstructions
\renewcommand{\confidential}{}
\renewcommand{\anonsubtext}{(No author info supplied here, for consistency with
TACL-submission anonymization requirements)}
\newcommand{\instr}
\fi

\iftaclpubformat %

\else

\fi

\title{Common Objects Out of Context (COOCo): Investigating Multimodal Context and Semantic Scene Violations in Referential Communication}

\author{
  Filippo Merlo\textsuperscript{1,2}, Ece Takmaz\textsuperscript{1}, Wenkai Chen\textsuperscript{1}, Albert Gatt\textsuperscript{1} \\ 
  \textsuperscript{1}Utrecht University \textsuperscript{2}University of Trento\\ 
  \texttt{f.merlo.research@gmail.com, e.k.takmaz@uu.nl} \\\texttt{w.chen5@students.uu.nl, a.gatt@uu.nl}
}

\date{}

\begin{document}
\maketitle
\begin{abstract}
To what degree and under what conditions do VLMs rely on scene context when generating references to objects?
To address this question, we introduce the \textit{Common Objects Out-of-Context (COOCo)} dataset and conduct experiments on several VLMs under different degrees of scene–object congruency and noise. We find that models leverage scene context adaptively, depending on scene-object semantic relatedness and noise level. Based on these consistent trends across models, we turn to the question of how VLM attention patterns change as a function of target-scene semantic fit, and to what degree these patterns are predictive of categorisation accuracy. We find that
successful object categorisation is associated with increased mid-layer attention to the target. We also find a non-monotonic dependency on semantic fit, with attention dropping at moderate fit and increasing for both low and high fit. These results suggest that VLMs dynamically balance local and contextual information for reference generation. 

Dataset and code are available here:
 \href{https://github.com/cs-nlp-uu/scenereg}{https://github.com/cs-nlp-uu/scenereg}.
\end{abstract}

\section{Introduction}
\label{intro}
\begin{figure}[t]
    \centering
    \begin{minipage}[h]{0.45\columnwidth}
    \includegraphics[width=\columnwidth]{images/000000064659_cubicle_office_laptop_original.jpg}
    \vspace{-0.8cm}
    \caption*{Original}
  \end{minipage}
  \hfill
  \begin{minipage}[h]{0.45\columnwidth}
    \includegraphics[width=\columnwidth]{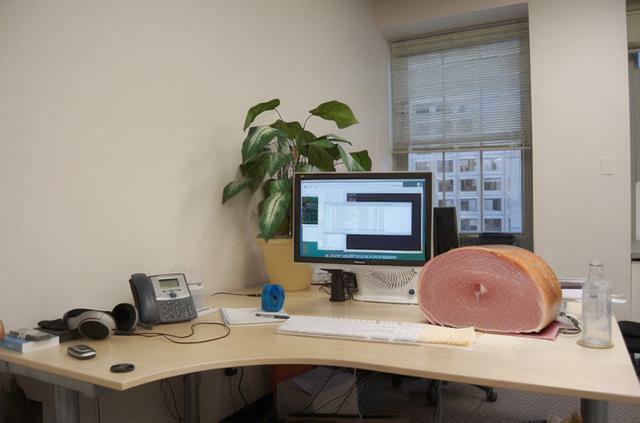}
    \vspace{-0.8cm}
    \caption*{Modified}
  \end{minipage}
   \hfill
  \begin{minipage}[h]{0.45\columnwidth}
    \includegraphics[width=\columnwidth]{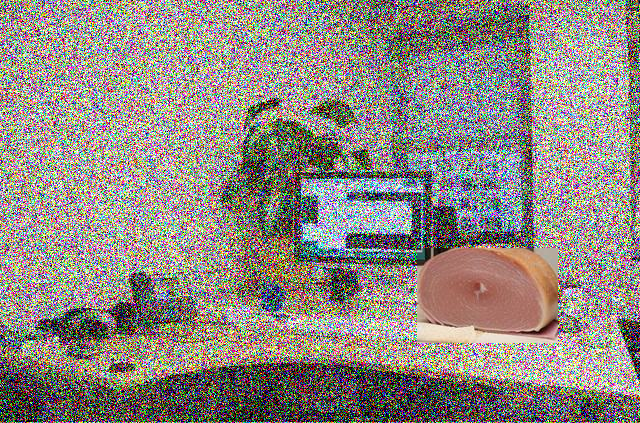}
    \vspace{-0.8cm}
    \caption*{Context Noise}
  \end{minipage}
   \hfill
  \begin{minipage}[h]{0.45\columnwidth}
    \includegraphics[width=\columnwidth]{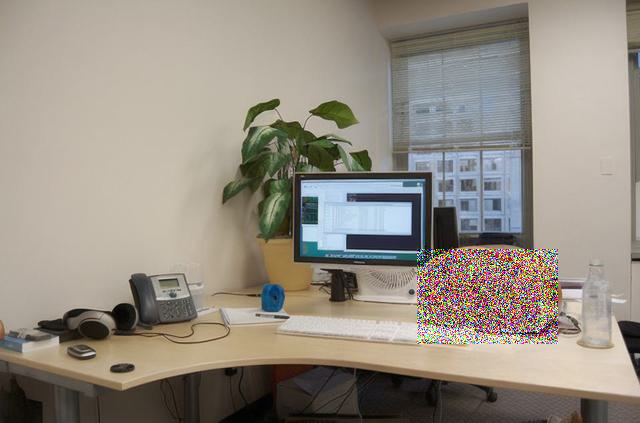}
    \vspace{-0.8cm}
    \caption*{Target Noise}
  \end{minipage}
    \caption{An original image of an office scene from the COCO dataset and a \textit{modified} version from COOCo, replacing the laptop with a semantically inconsistent object. Bottom: versions of the modified image with Gaussian noise mask (level = 0.5) on the context and on the target.} %
    \label{fig:ham_office}
\end{figure}

Objects rarely appear in isolation. Over time, humans build expectations of seeing certain objects in certain contexts.
For instance, we expect to see a laptop on a desk in an office, but not a large piece of ham (Figure~\ref{fig:ham_office}). Studies of human scene perception~\cite{vo_meaning_2021, Torralba2006-TORCGO, Coco2016-COCAIR} suggest that scenes exhibit both semantic and syntactic regularities~\cite{Biederman1982,vo_meaning_2021}. Violations of either or both of these result in processing difficulties ~\cite{Ganis2003,SCEGRAMAI,vo_meaning_2021}.

In this work, we are interested in how Vision-Language Models (VLMs) behave in the presence of \emph{semantic} violations in visual scenes when referring to objects. 
We focus on {\bf object naming}, that is, the task of determining the category of an object, a fundamental part of Referring Expression Generation~\cite[REG;][]{krahmer_computational_2012}.
The literature on scene perception suggests that scene semantics mediates human viewers' expectations about the objects in a scene, and viewers deploy their attentional resources accordingly. Scene semantics also influence how humans {\em describe} objects in scenes~\cite{Hwang2011,de_groot_language-induced_2017,DeGroot2015}. Do VLMs similarly utilise scene-level visual information when generating descriptions of objects? To study this, we compare the naming of objects in semantically (in)congruent scenes. If scene-level information guides model choices, then there should be systematic differences between these two conditions in the way models refer to target referents. While previous work has shown that visual context has a facilitating effect in REG~\citep{junker-zarriess-2024-resilience-scene}, it does not control for the impact of semantic (in)congruity, and relies on small-scale, purposely-trained models or on artificial images~\citep{junker_scenegram_2025}. 
To address these issues, we develop a new dataset, Common Objects Out-of-Context (\textbf{COOCo}), which manipulates natural images to introduce objects of 
varying degrees of consistency with the scene (Figure~\ref{fig:ham_office}). Similarly to \citet{junker-zarriess-2024-resilience-scene}, we further manipulate the visibility of the target referent and the context by introducing different degrees of noise. Furthermore, since different objects can be related to a scene to different degrees, we model object-scene relatedness as a graded phenomenon.
We summarise the contributions of our study as follows.

\noindent
{\bf 1.} We present COOCo, a novel dataset to evaluate multimodal models' ability to integrate object-level and scene-level visual information, under both congruent and incongruent conditions. 

\noindent
{\bf 2.} We investigate the impact of scene context on object identification on several state-of-the-art VLMs. We show that under conditions where target objects violate scene semantics, scene context acts as a distractor, while it has a clear facilitating effect when the target is congruent with the scene. This also makes models resilient to noise when it is applied to the target region.

\noindent
{\bf 3.} Having established that object-scene relatedness exerts an effect across models, we then focus on a single, representative VLM. Using COOCo and another, recently introduced dataset, we investigate how it deploys attention across scene components under different conditions.

\section{Related Work}
\label{relwork}
\paragraph{Context in Referring Expression Generation}
Referring Expression Generation (REG) is the task of generating identifying descriptions for a target object. Early REG approaches relying on handcrafted algorithms \cite{krahmer_computational_2012} incorporated contextual influences by considering the properties that distinguish a target referent from its distractors, with some research also incorporating constraints to guide coherent categorisation for references to plurals, and the use of landmarks for spatial expressions
\cite{schuz_rethinking_2023}. 
More recent neural models for \emph{visual REG} learn mappings from raw perceptual inputs to linguistic expressions. Here too there has been limited attention on the ability of models to leverage global scene information \cite{schuz_rethinking_2023}. This gap has motivated recent investigations into 
how scene information may enhance robustness in challenging settings \cite{schuz_keeping_2023, junker_scenegram_2025}.
In particular, \citet{junker-zarriess-2024-resilience-scene} show that models rely on scene context to generate references even when the target object is fully occluded, indicating that scene context serves both a supportive and disambiguating role during reference generation. \citet{junker_scenegram_2025} further find that scene context influences how VLMs describe ambiguous tangram shapes, though to a lesser extent than humans.

\paragraph{Context in Human Object Recognition} 
Human object recognition is strongly influenced by contextual expectations
\cite{bar_visual_2004, oliva_role_2007, vo_meaning_2021, peelen_predictive_2023}, including both global scene properties (e.g., gist, layout) and local cues (e.g., anchor objects, co-occurring items) 
\cite{lauer_role_2018, lauer_role_2021, boettcher_anchoring_2018}. 
Evidence that human overt attention in visual search is semantically constrained comes from viewing tasks, where viewers fixate more on semantically informative (but not necessarily visually salient) regions,
and semantic similarity is a strong predictor of gaze trajectories \cite{henderson_meaning_2019, hayes_looking_nodate}. Both highly expected and highly incongruent objects attract more attention than moderately fitting ones \cite{damiano_exploring_2024}. Motivated by this work, we take a graded view of object-scene semantic similarity in the design of our dataset.

\paragraph{Datasets}
Widely used vision and language datasets for referring expression and image caption generation, such as COCO  \cite{lin_microsoft_2015-1}, RefCOCO \cite{kazemzadeh_referitgame_2014, Yu2016} and VisualGenome \cite{krishna_visual_2016}, lack annotations for object-scene relatedness.
Scene violation datasets created within the vision community include
ObjAct \cite{shir_you_2021}, Event Task \cite{dresang_semantic_2019}, Out-of-Context \cite{bomatter_when_2021}, Cut-and-paste dataset \cite{yun_cut-and-paste_2021}, and SCEGRAM \cite{SCEGRAMAI}. 
However, these datasets tend to be relatively small-scale and do not incorporate degrees of semantic violation or occlusion.
In this work, we present a new, large-scale dataset for semantic violations in visual scenes, building on existing resources  
such as COCO-Search18 \cite{chen_coco-search18_2021}, THINGSPlus \cite{hebart_things_2019, stoinski_thingsplus_2023} and SUN \cite{xiao_sun_2010}. Most related to our work are two recent and concurrently released datasets: VISIONS~\cite[][further described in \S~\ref{dataset}]{allegretti_visual_2025} incorporates semantic and spatial violations but remains small in scale (1,136 images), providing a single incongruent manipulation per scene. COinCO~\cite{yang_common_2025}
introduces one inpainted object replacement per image and adopts a binary notion of contextual fit based on judgements generated with Molmo \cite{deitke2024molmopixmoopenweights}.
In contrast, COOCo offers multiple controlled variants of each scene by systematically modulating object–scene relatedness on a continuous scale, enabling finer semantic manipulation than prior datasets.
We also evaluate on VISIONS because it is fully human-validated, provides a continuous object-scene relatedness measure, and includes modal object names.

\section{Dataset Creation}
\label{dataset}
We introduce the \textbf{Common Objects Out-of-Context (COOCo)} dataset.
The dataset provides a collection of images where the semantic coherence of a scene is modified by the presence of a target object with low, medium or high semantic relatedness to the scene type. 
COOCo builds upon a framework previously established in the visual perception literature with the SCEGRAM dataset \cite{ohlschlager_scegram_2017}, scaling it up to meet the needs of NLP and computer vision research through a novel methodology.

\paragraph{Images} 
We begin with the publicly available subset of COCO-Search18 \cite{chen_coco-search18_2021}, which includes bounding boxes for target objects.
This dataset excludes images with people or animals, categories which are known to significantly influence human visual attention
~\cite{Clarke2013a,end_preferential_2017}. Excluding such cues, therefore, avoids confounding factors and enables us to study object-scene semantic relationships. COCO-Search18 further excludes images with targets with multiple object instances, controlling for target size, location, and image aspect ratio. We retained the 2,241 images where the target object is present. 

\paragraph{Scene labels}
For each image, we predict the scene label using a Vision Transformer (ViT)\footnote{https://huggingface.co/google/vit-base-patch16-224} \cite{dosovitskiy_image_2021} finetuned on SUN-397 \cite{xiao_sun_2010}.
We then reduce the SUN label set (397 labels) to 29 (listed in Appendix~\ref{app:scenes}),
selected to balance frequency and semantic diversity and reclassified the images with the same model, excluding all but the selected labels.
This approach mitigates risks of category under-representation and excessive variance. This step provided a scene label for each image. 

\begin{figure*}[h!]
    \centering
    \setlength{\tabcolsep}{1em} %
    \renewcommand{\arraystretch}{5} %
    \begin{tabular}{ccc} %
        \begin{minipage}{0.75\textwidth}
            \centering
            \parbox{0.28\textwidth}{\centering \quad\textbf{Original} \\ \includegraphics[width=0.32\textwidth]{images/000000064659_cubicle_office_laptop_original.jpg}}
            \hfill
            \parbox{0.28\textwidth}{\centering \quad\textbf{Clean} \\ \includegraphics[width=0.32\textwidth]{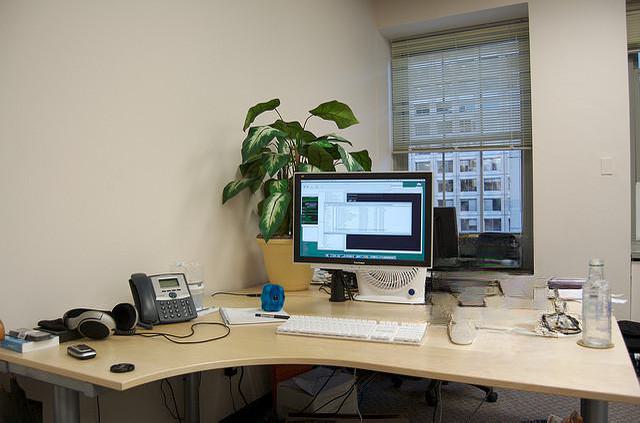}}
            \hfill
            \parbox{0.28\textwidth}{\centering \quad\textbf{Generated} \\ \includegraphics[width=0.32\textwidth]{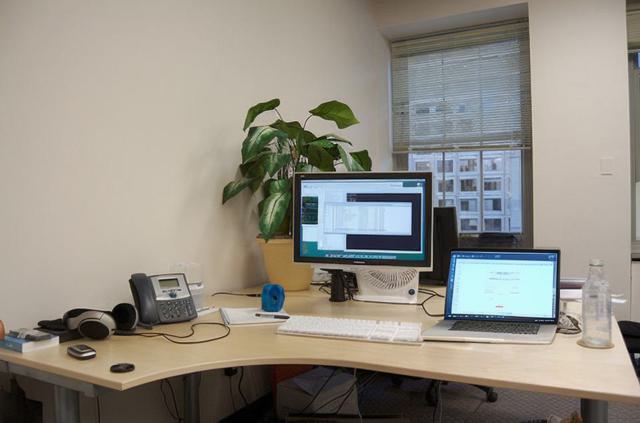}}
        \end{minipage}\\
 
        \begin{minipage}{0.75\textwidth}
            \centering
            \vspace{0.5cm}
            \parbox{0.28\textwidth}{\centering \quad\textbf{High} \\ \includegraphics[width=0.32\textwidth]{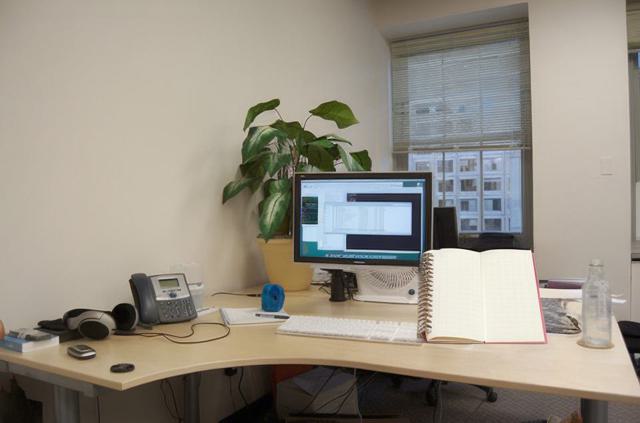}}
            \hfill
            \parbox{0.28\textwidth}{\centering \quad\textbf{Medium} \\ \includegraphics[width=0.32\textwidth]{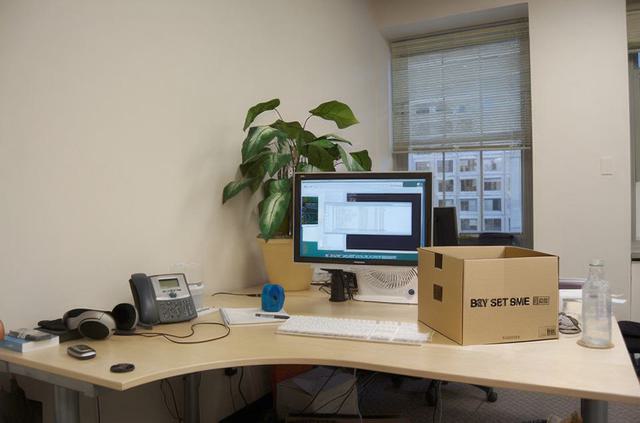}}
            \hfill
            \parbox{0.28\textwidth}{\centering \quad\textbf{Low} \\ \includegraphics[width=0.32\textwidth]{images/000000064659_cubicle_office_laptop_ham_relscore_low.jpg}}
        \end{minipage}
    \end{tabular}
    
    \caption{A set of \emph{COOCo} images from the "cubicle office" scene category with target object {\em laptop}. The target is removed from the original image (`clean') and replaced with objects of the same type (`generated'), as well as targets with high-, low- and medium relatedness to the scene. 
    }
    \label{fig:relatedness_scores}
\end{figure*}

\paragraph{Object-scene congruency} 
We developed an inpainting pipeline to create versions of images where the target object is replaced with alternatives of varying congruence with the scene (low and medium).
We also retain the original image, and additionally include controls where (a) the target is replaced with an object that has high relatedness to the scene, and (b) the target is replaced with a new object of the same category as the original. The latter are included to verify if the effects found in the experiments are due to artefacts caused by inpainting in general, or due to the introduction of incongruent object types. 
Figure~\ref{fig:relatedness_scores} shows an example of a scene and its variants in COOCo.

\noindent
New objects are determined using human-generated norms in THINGSplus
\cite{stoinski_thingsplus_2023}, 
which includes typicality ratings (the degree of representativeness of an object for higher-level categories) and ratings of the real-world size of objects. We retain only typical objects (typicality ranging from 0.3 to 1) and remove animate objects to ensure consistency with COCO-Search18.
We then match COCO-Search18 targets to candidate objects in THINGSPlus with similar real-world size norms, excluding any candidates differing in size from the targets by more than 25 units.

\noindent
We compute semantic relatedness between candidate replacements and scene labels following \citet{hayes_looking_nodate}, who found strong correlations between embeddings in ConceptNet Numberbatch~\cite{speer_conceptnet_2018} and human visual attention shifts. We use the cosine similarity between scene and object label embeddings to filter low-, medium- and high-similarity candidates.

\paragraph{Image modification}
Given an image, a scene label and a target object with its bounding region and  
segmentation-based mask obtained from COCO-Search18,
we remove the target from the image, and generate
15 new versions of the scene where the target is replaced with objects having low, medium or high similarity to the scene label (see Figure~\ref{fig:relatedness_scores}). We also generate an image containing the target as in the original image as a control condition. Only images that pass a verification test are retained. Full details of the inpainting pipeline and verification tests are in Appendix~\ref{app:scenes}.

\paragraph{Final COOCo dataset} The final COOCo dataset comprises 18,395 images: 1,862 original images with corresponding object-removed versions; 5,480 generated images with low relatedness; 5,572 with medium relatedness; 1,771 with high  relatedness; and 1,848 images for the same-target condition. 

\paragraph{VISIONS dataset} 
We conduct a subset of our experiments on the VISIONS dataset \cite{allegretti_visual_2025} and compare the results obtained on COOCo with those on it. VISIONS comprises 1,136 naturalistic indoor images constructed from 165 scene setups across seven scene categories: bathroom, bedroom, church, kitchen, office, restaurant, and waiting room. Each setup is photographed in four principal versions created by manipulating the semantic consistency (consistent vs. inconsistent) of a target object, identified with bounding box coordinates, and its spatial position (left vs. right).\footnote{Although the dataset also includes additional swap-based variants for 119 scenes where the target object was exchanged with another object in the scene to produce four mirrored counterparts of the canonical versions, we use only these four canonical versions, as swap manipulations are not part of our analyses.}
VISIONS provides an extensive set of perceptual and conceptual norms for both objects and scenes, including 
object-consistency ratings, which quantify how contextually appropriate the critical object is in each scene. We utilise these ratings. VISIONS also provides human-annotated target objects with free-text labels; we use the modal name for each object as the reference label in our analysis, together with the original object label defined by the dataset creators.

\section{Experimental Setup}
\label{exps}
We design our experiments to address how VLMs leverage contextual information for object naming, an essential component of Referring Expression Generation. (1) First, we evaluate to what extent VLMs rely on scene context under conditions of varying degrees of semantic congruence between an object and its background; and on how robustly they perform REG under different levels and configurations of visual noise. These experiments are conducted on a variety of SOTA VLMs on the COOCo dataset. Each model is provided with an image and a specified target region, and is prompted to name the object within that area using a structured text prompt. Anticipating the outcomes of this analysis (see \S~\ref{results}), we identify a number of trends across models, showing that they are sensitive to scene semantics when referring to targets and that scene context supports REG when target objects are obscured by noise \cite[cf.][]{junker-zarriess-2024-resilience-scene}. 
(2) Having established these trends across multiple models, we select a single performant model for in-depth analysis of its outputs (\S~\ref{res_out_vs_scene}) and its attention deployment across different experimental settings (\S~\ref{exp_attention}). This analysis is performed on our own dataset, COOCo, with replications on the VISIONS dataset \cite{allegretti_visual_2025}. This allows us both to generalise findings and control for possible artefacts introduced by inpainting in COOCo.

\subsection{Models}
\label{exp_models}
We use the following models in our experiments:

\noindent
{\bf KOSMOS-2}
\cite{peng2024grounding} (1.6B) 
incorporates mechanisms to establish direct associations between textual spans and specific image regions. It
is trained on the GrIT (Grounded Image-Text pairs) web-scale dataset.

\noindent
{\bf Molmo}
\cite{deitke2024molmopixmoopenweights} (7B) is trained on a dataset including ca. 2.3M human-annotated pairs of questions and references to image regions based on 2D points, from 428k images.

\noindent
{\bf xGen-MM-Phi3/BLIP-3}
\cite{xue2024xgenmmblip3familyopen} (ca. 4.4B)  is trained on the BLIP3-GROUNDING-50M dataset and
supports spatial encoding using bounding box coordinates, descriptive spatial relations 
and/or relative positioning.

\noindent
\textbf{LLaVA-OneVision}
\cite{li2025llavaonevision} (0.5B and 7B) was trained on data with bounding boxes, incorporating RefCOCO \cite{kazemzadeh_referitgame_2014, Yu2016}[>50k samples] and Visual Genome \cite{krishna_visual_2016}[86k] for single-image settings, and multi-image datasets including WebQA \cite{chang_webqa_2022}[9.3k]. The model is instruction-tuned to accept spatial information directly in text input in the form of normalised bounding box coordinates.

\noindent
{\bf Qwen2.5-VL}
\cite{bai_qwen25-vl_2025} (7B)
is trained with coordinates based on images' native resolution for grounding, detection,
pointing and object counting. 
The model accepts absolute coordinates for image regions as part of its prompt. 

For detailed descriptions of preprocessing steps and region-of-interest (ROI) prompt formats for each model, refer to Appendix~\ref{app:preproc}.

\paragraph{Prompts} We test models with structured prompts (details in Appendix~\ref{app:preproc}) with an explicit reference to a region, for example: {\em What is the object in this part of the image [$x_1$, $y_1$, $x_2$, $y_2$]? Answer with the object's name only. No extra text.}

\subsection{Noise injection}
We add Gaussian noise (\( \lambda = 0.0, 0.5, 1.0 \)), ranging from mild distortion to maximum information loss, to the target region, the context region or the entire image (`All noise'). 
Pixel values in the affected area are perturbed with noise sampled from a normal distribution and clipped to stay within valid image ranges. 

\subsection{Metrics}\label{metrics}
We evaluate models using three metrics: 

\noindent
\textbf{RefCLIPScore} \cite{hessel_clipscore_2022} is the harmonic mean
between (1) the \texttt{CLIPScore} \cite{hessel_clipscore_2022}, which captures the semantic alignment between the generated caption and the image; and (2) the highest cosine similarity between the CLIP~\cite{radford_learning_2021} embeddings of the generated caption and any reference caption. This quantifies the alignment between a generated description for a target and both its visual features (extracted from its image patch) and correct label.

\noindent
{\bf Text-Based Semantic Similarity} is computed as the cosine similarity between CLIP text embeddings. We use this measure to assess the semantic similarity between the model’s output and the target label (from which Accuracy is derived), as well as between the model’s output and the scene label in \S~\ref{res_out_vs_scene}.

\noindent
\textbf{Accuracy}, defined as 1 if the cosine similarity between the model's output and the target label exceeds a threshold (0.9), and 0 otherwise. To illustrate, consider the target \emph{car}. While the output \emph{car} scores 1, related terms \emph{bus} (0.86), \emph{truck} (0.89), and \emph{police car} (0.89) score 0. In contrast, variants such as \emph{SUV} (0.90), \emph{a race car} (0.90), and the part–whole expression \emph{car door} (0.90) surpass 0.9 and are counted as correct.\footnote{We also computed a hard accuracy metric based on Levenshtein similarity, which quantifies the character-level edit distance between generated and reference captions. 
Since results are very similar to those obtained with the cosine-based metric, we do not report them here.}

\section{Results across models}
\label{results}
In this section, we report and discuss the main results across all models and experimental conditions in COOCo. 

\begin{figure*}[h]
    \centering
    \includegraphics[scale=0.25]{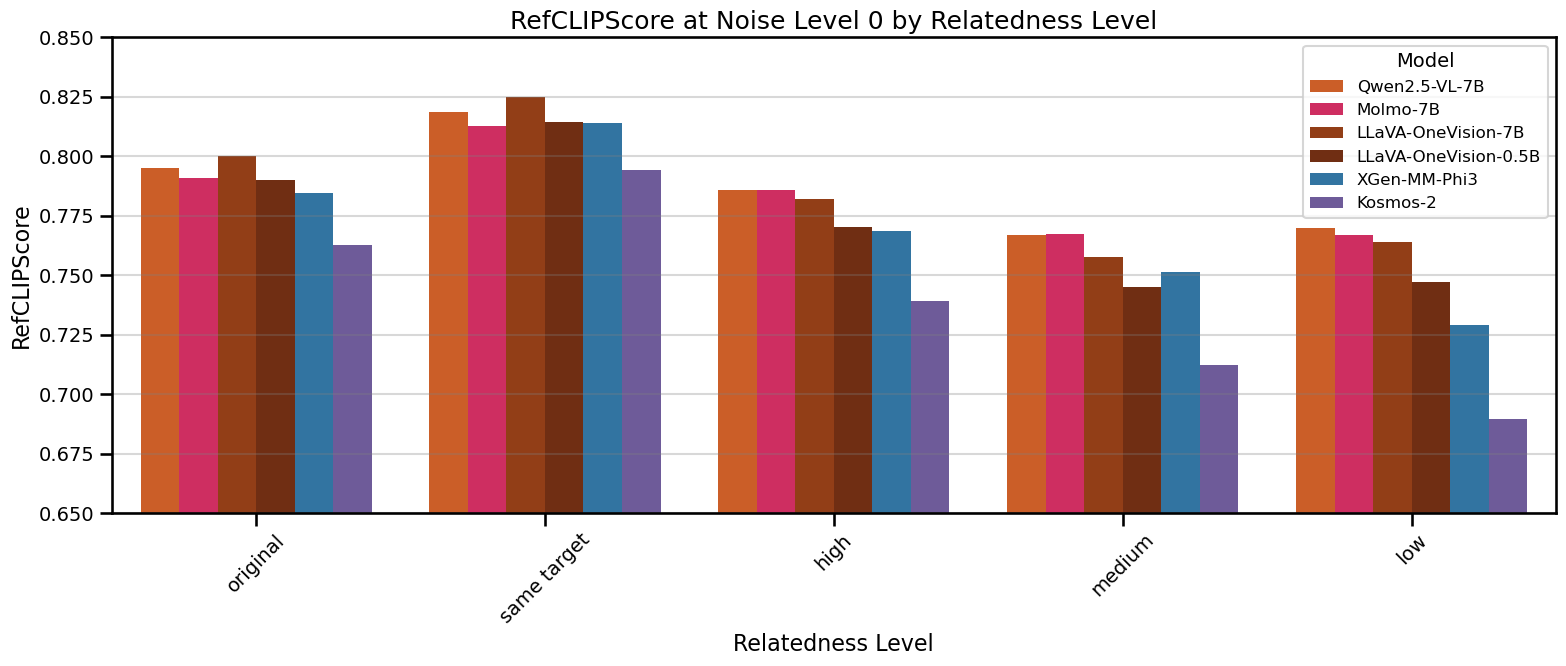}
    \caption{RefCLIPScores per model at noise level 0 across relatedness conditions. Models perform best in the original and same target conditions, with performance declining as target relatedness decreases. }
    \label{fig:refclipscore_sing_models}
\end{figure*}

\begin{figure*}[h]
    \centering
    \begin{subfigure}{0.45\textwidth}
    \centering
    \includegraphics[scale=0.3]{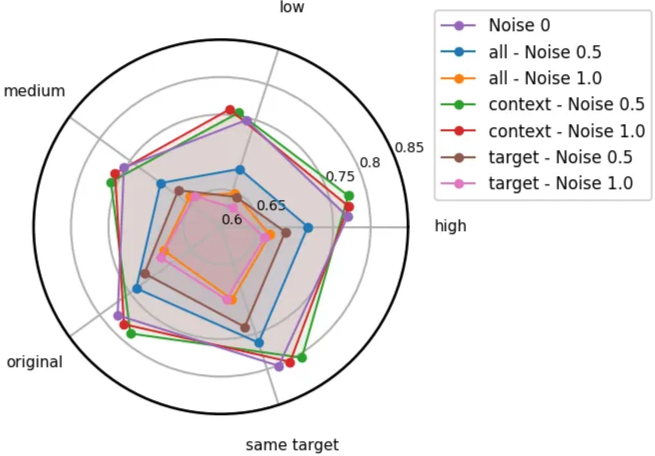}
    \caption{RefCLIPScores by relatedness, noise area, and noise level.}
    \label{fig:refclipscore_aggregated}
    \end{subfigure}
    \hfill
    \begin{subfigure}{0.45\textwidth}
    \centering
    \includegraphics[scale=0.3]{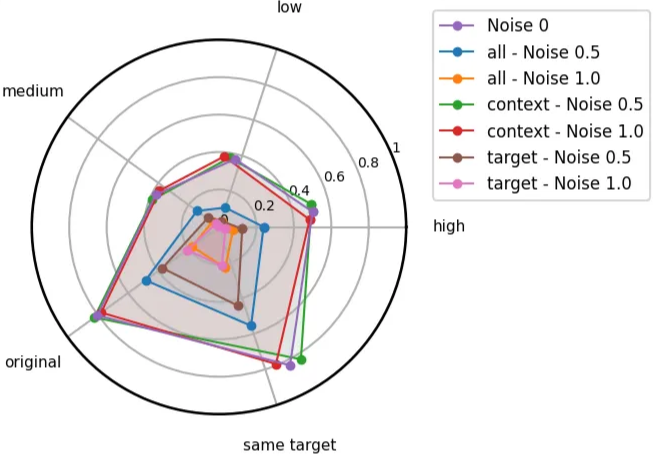}
    \caption{Accuracy across noise levels, noise areas, and relatedness conditions. }
    \label{fig:soft_acc}
    \end{subfigure}
    \caption{RefCLIPScore and Accuracy across experimental conditions, aggregated over all models. When differences within a condition are minimal, the leftmost point corresponds to the higher value, followed by lower values to the right.}
\end{figure*}

\paragraph{Models are sensitive to scene semantics when referring to targets.}
Figure~\ref{fig:refclipscore_sing_models} reports RefCLIPScore results for each model at noise level 0 across all relatedness conditions. Models obtain the highest scores in the {\em original} and {\em same target} conditions, indicating stronger alignment with the reference when the target object is preserved or replaced by an object from the same category. Interestingly, performance is higher in the same target condition than in the original. A possible reason for this is that inpainting introduces subtle artefacts that make it easier for target objects to be picked out. Performance decreases when the target object differs from the original ({\em high}, {\em medium}, and {\em low} relatedness). Within these conditions, models perform best under high relatedness, with scores progressively declining as relatedness decreases. Importantly, the observed effects of object–scene relatedness are also supported by results on VISIONS, where no inpainting is applied (see \S~\ref{sec:llava}).

For models depicted in warm colours (the first four in Figure~\ref{fig:refclipscore_sing_models}), the low relatedness condition yields performance that is similar to, or slightly higher than, the medium condition. Conversely, models in cool colours (XGen-MM-Phi3 and Kosmos-2) show a more consistent decline in performance with decreasing relatedness.

\paragraph{Scene context acts as a distractor for targets that violate scene semantics.}
Figure~\ref{fig:refclipscore_aggregated} reports RefCLIPScore aggregated across models by relatedness, noise area, and noise level. Overall, models achieve their best performance at context noise 0.5, followed by noise 1.0 and then the zero-noise condition, with the notable exception of the low-relatedness condition, where noise 1.0 yields the highest alignment. Consistent trends are observed in the accuracy results (Figure~\ref{fig:soft_acc}): the {\em original} and {\em same target} conditions reach the highest accuracy ($\sim$0.8) across noise settings, followed by the high-relatedness condition ($\sim$0.5) and the medium and low-relatedness conditions ($\sim$0.4). While most relatedness conditions peak at context noise 0.5, the low-relatedness condition shows a monotonic improvement with increasing noise, indicating that maximal removal of contextual information is particularly beneficial when targets violate scene semantics and scene context becomes detrimental.

\paragraph{Scene context acts as a facilitator when congruent targets are obscured.}
In contrast, applying noise directly to the target region results in a clear performance degradation, proportional to the noise intensity: the higher the noise level, the greater the drop. When noise (at 0.5) is applied to the target or the entire image, accuracy in the original and same target conditions drops to ~0.5, while all other conditions approach zero (Figure~\ref{fig:soft_acc}).
For RefCLIPScore, we find that at 0.5 noise level, for the {\em high}, \textit{medium}, and especially \textit{low} relatedness conditions, performance is more robust when noise is applied to the entire image rather than solely to the target. At a higher noise level (1.0), performance continues to decline. However, 
models perform best with target-specific noise in the \textit{original} and \textit{same target} conditions. This suggests that, given a non-perturbed scene context, model predictions for noised target regions are for objects that would be expected given the scene composition (see \S~\ref{res_out_vs_scene}).
Comparing the 0.5 and 1.0 all-noise settings, we see a sharper drop for the original and same-target conditions, while the drop is smaller for the high, medium and especially the low relatedness conditions.

\paragraph{Mixed-Effects Analysis of RefCLIP Similarity}
Focusing on the `extreme' semantic conditions ({\em original} and {\em low} relatedness), we build a Linear Mixed-Effects (LME) model with RefCLIPScore as the dependent variable, noise level (scaled and centred) as a continuous predictor, and noise area as a categorical factor contrasting context and target noise against the {\em all-noise} baseline. The model includes random intercepts for the image identifier and the VL models.\footnote{Since we are interested here in establishing the statistical reliability of the trends observed above across all models.} 
Table~\ref{tab:lme_refclipscore} reports the LME results. 
Confirming our observations, baseline RefCLIPScore is higher for the original than for the low-relatedness images. Context noise increases similarity, while target noise decreases it. As expected, this contrast is stronger in the low-relatedness condition.
Noise intensity shows a significant negative main effect, with interactions showing a less steep negative impact when noise is in the context or the target compared to full-image noise.

\begin{table*}[!t]
\centering
\footnotesize
\begin{tabular}{l|ccc|ccc}
\toprule
 & \multicolumn{3}{c|}{\textbf{Original images}} 
 & \multicolumn{3}{c}{\textbf{Low relatedness}}\\
\textbf{Predictor} 
 & \textbf{Coef.} & \textbf{Std.Err.} & \textbf{z}
 & \textbf{Coef.} & \textbf{Std.Err.} & \textbf{z} \\
\midrule

Intercept  
 & 0.735 & 0.002 & 379.21$^{*}$
 & 0.692 & 0.003 & 275.51$^{*}$ \\

Noise area (context)
 & 0.052 & 0.001 & 94.13$^{*}$
 & 0.062 & 0.000 & 141.32$^{*}$ \\

Noise area (target)
 & -0.015 & 0.001 & -27.34$^{*}$
 & -0.047 & 0.000 & -107.65$^{*}$ \\

Noise level (cont.)
 & -0.034 & 0.000 & -145.42$^{*}$
 & -0.034 & 0.000 & -182.28$^{*}$ \\

Noise area (context) × Noise level
 & 0.032 & 0.001 & 60.75$^{*}$
 & 0.036 & 0.000 & 86.79$^{*}$ \\

Noise area (target) × Noise level
 & 0.014 & 0.001 & 26.76$^{*}$
 & 0.024 & 0.000 & 56.84$^{*}$ \\
 
Group Var
 & 0.041 &   &  
 & 0.069 &   &   \\

\bottomrule
\end{tabular}
\caption{
Linear mixed effects model: Noise area and noise level on RefCLIPScores, for original and low-relatedness images. $^*$ denotes a significant effect at $p \leq .001$.
}
\label{tab:lme_refclipscore}
\end{table*}

\begin{figure*}[!t]
    \centering
    \includegraphics[width=\linewidth, trim=0 0 0 0, clip]{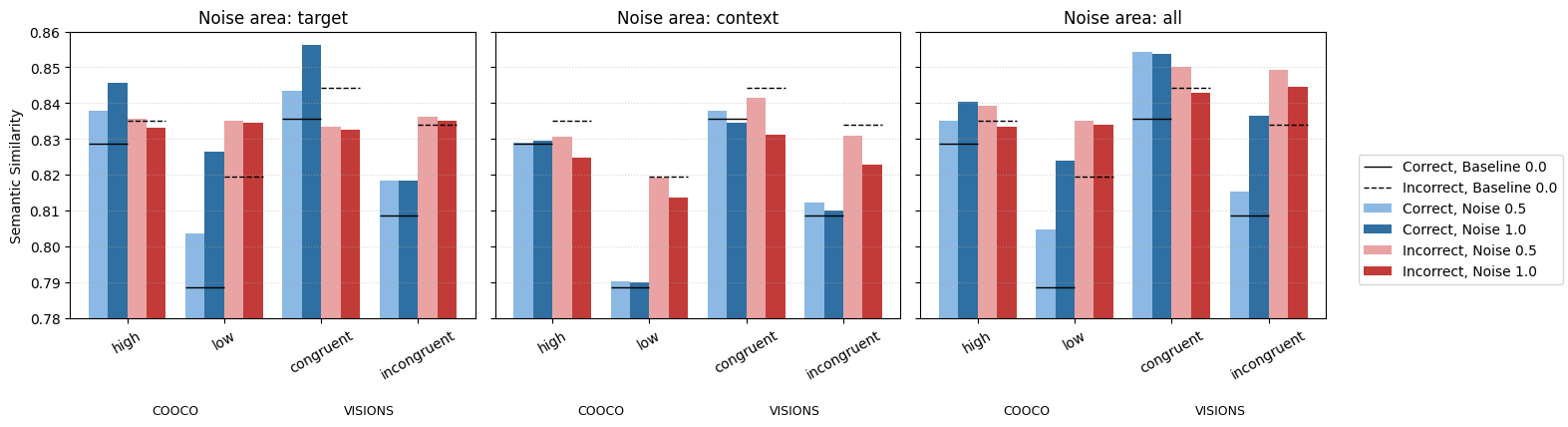}
    \caption{Semantic similarity between the model’s outputs and scene labels for LLaVA-OneVision-0.5B in COOCo (ours) and VISIONS~\cite{allegretti_visual_2025}, shown across correctness (blue: correct; pink/red: incorrect), semantic-fit conditions (COOCo: high/low; VISIONS: congruent/incongruent), noise areas (target, context, all), and noise levels (0.5, 1.0). Dashed and solid horizontal lines indicate baseline similarity at zero noise for incorrect and correct predictions, respectively.}
    \label{fig:scene_vs_output_comp}
\end{figure*}

\begin{table*}
\centering
\footnotesize
\begin{tabular}{l|ccc|ccc}
\toprule
    & \multicolumn{3}{c|}{\textbf{Original images}} & \multicolumn{3}{c}{\textbf{Low relatedness}}\\
 & \textbf{Coef.} & \textbf{Std.Err.} & \textbf{z} & \textbf{Coef.} & \textbf{Std.Err.} & \textbf{z} \\
\midrule
Intercept & 1.639 & 0.044 & 37.538$^*$ & 1.598 & 0.034 & 46.716$^*$ \\
Noise: Context & -0.006 & 0.007 & -0.787 & -0.183 & 0.004 & -41.733$^*$ \\
Noise: Target & -0.048 & 0.007 & -7.364$^*$ & -0.032 & 0.004 & -8.042$^*$ \\
Noise level (cont.) & -0.068 & 0.007 & -10.377$^*$ & -0.004 & 0.004 & -1.062 \\
Noise level $\times$ Noise: Context & 0.051 & 0.009 & 5.472$^*$ & -0.016 & 0.006 & -2.867$^*$ \\
Noise level $\times$ Noise: Target & 0.039 & 0.007 & 5.364$^*$ & 0.066 & 0.004 & 14.865$^*$ \\
Group Var& 0.047 & 0.048 &  & 0.029 & 0.029 &  \\
\bottomrule
\end{tabular}
\caption{Linear mixed effects model: Noise area and noise level on scene-target semantic similarity, for original and low relatedness contexts. $^*$ denotes a significant effect at $p \leq .01$. Conditional $R^2$ values: 0.52 (original) and 0.39 (low rel.)}
\label{tab:lme}
\end{table*}

\section{In-depth analysis of LLaVA-OneVision}\label{sec:llava}
We follow up with further analysis of LLaVA-OneVision-0.5B, chosen because of its size, which makes in-depth exploration feasible and its competitive performance (cf. Figure~\ref{fig:refclipscore_sing_models}; full accuracy scores for this model are in Appendix~\ref{app:results}, Tab~\ref{tab:cooco-acc} and \ref{tab:visions-acc}). We report analyses on both COOCo and VISIONS~\cite{allegretti_visual_2025}, rescaling images in the latter from the original (1920 $\times$ 1080 px) to match the resolution in COOCo (max. width of 640 px). We note that for VISIONS, naming differences between dataset labels and human-provided modal object names have negligible impact on accuracy (see Appendix~\ref{app:modal_names}).

\subsection{Output vs. scene similarity}\label{res_out_vs_scene}
We begin with an analysis of the text-based semantic similarity metric scores between the model's predicted reference to the target (e.g. \textit{laptop}), and the scene label for an image (e.g. \textit{office}),
to evaluate whether the generation was more strongly anchored to the overall scene or to the target object itself.
Figure \ref{fig:scene_vs_output_comp} reports how semantic similarity between the model’s outputs and the scene labels varies across correctness, noise conditions and intensity, and the relatedness distinctions defined in each dataset: high vs. low relatedness in COOCo, and congruent vs. incongruent scenes in VISIONS.\footnote{
Appendix~\ref{app:results}, Fig.~\ref{fig:scene_vs_output_continuous} presents a complementary analysis, treating object-scene relatedness as continuous rather than categorical.}

When the target region contains \emph{no noise} (in the baseline condition with 0 noise, and in the context-noise setup), the same trend appears in both correct and incorrect outputs: in low-relatedness conditions, the model produces predictions that are less semantically aligned with the scene label than in the higher-relatedness conditions of each dataset. This indicates that, when the target object violates scene semantics, predictions are less anchored to the global scene and are instead more influenced by the unrelated object itself. Crucially, this pattern disappears when noise is applied to the \emph{target region}. Incorrect predictions under target noise no longer exhibit clear differences between high and low semantic fit, suggesting that once the target's appearance is degraded, object semantics cease to guide outputs. In contrast, correct predictions under target noise consistently shift towards the scene label: across all semantic-fit conditions, correct responses exhibit higher similarity to the scene than in the no-noise baseline (solid line). This aligns with the observation in \S~\ref{results} that degrading the target prompts the model to compensate by relying more heavily on global scene context. When successful, the resulting predictions are more semantically aligned with the scene. Notably, only in the congruent case of the VISIONS dataset, incorrect predictions fall below their zero-noise baseline, suggesting that congruent objects, when not degraded, may anchor the answers to a higher semantic similarity with the scene. 

When noise is applied to the \emph{context region}, correct responses exhibit no meaningful deviation from the zero-noise baseline, indicating that when the target remains intact, the model continues to anchor its prediction primarily to the object itself rather than the broader scene. Incorrect responses, however, show a general decrease in similarity relative to the baseline, suggesting that context degradation harms scene-based anchoring when the model fails to identify the target correctly.
Under \emph{all noise} conditions, the effects differ across datasets: COOCo exhibits patterns comparable to those observed under target noise alone, whereas in VISIONS, the incorrect predictions in congruent trials maintain their baseline similarity to the scene, and those in incongruent trials even show an increase in similarity.

We further analyse these results using LME on the COOCo data; the same analysis on VISIONS is reported in Appendix~\ref{app:results}, Tab~\ref{tab:lme-visions}. 
We treat semantic similarity (logit-transformed) between predicted target and scene label as the dependent variable. Noise level (modeled as a continuous variable, scaled and centred) and area (comparing target and context to the {\em all} noise condition) are predictors, with random intercepts for scene labels.\footnote{We do not include a full random effects structure, as introducing random slopes resulted in poor convergence.} We include both correct and incorrect outputs, analysing separately the two `extreme' cases of target-scene congruence, namely, the {\em original} and the {\em low-relatedness} conditions. 
Table~\ref{tab:lme} summarises the LME results. The main effect of noise level is significant in the {\em original}, but not the {\em low relatedness}condition. The {\em target} noise area is signficantly different from the {\em all} noise condition, but {\em context} is only significant in low-relatedness images. (All main effects are significant in VISIONS; cf. Tab.~\ref{tab:lme-visions}). More important are the interactions: In original images, increased noise in the target or context region results in slight but significant increases in semantic similarity of the predicted target to the scene label. In the low relatedness condition, increasing noise in the context region results in a significant decrease in similarity between model predictions and scene label.

\subsection{Attention Deployment Analysis}\label{exp_attention}
We now turn to an analysis of the model's attention dynamics under different experimental conditions. In what follows, we use the term `attention' to refer to the Transformer decoder's attention to the image tokens when it generates the name of the object.
To analyze attention allocation across different layers \( l \) of the vision encoder and image samples \( t \), we compute the ratio \( r \) between the total normalized attention values assigned to the target (\( a_{\mathrm{target}} \)) and the context (\( a_{\mathrm{context}} \)): $r_l^{(t)} = \frac{a_{\mathrm{target},l}^{(t)}}{a_{\mathrm{context},l}^{(t)}}$.
A value of \( r \) higher than 1 indicates a higher allocation of attention to the target region, whereas a value closer to 0 suggests greater attention to the context. Since the context area is larger than the target, we do not expect \( r \) to exceed 0.5. Instead, our focus is on analysing its variations across different experimental conditions.
We then compute the mean ratio value for each encoder layer by averaging across all samples: $\bar{r}_l = \frac{1}{T} \sum_{t=1}^{T} r_l^{(t)}$
where \( T \) denotes the total number of instances. This allows us to assess how attention is distributed across different hierarchical levels of the encoder under varying experimental conditions. 
Full details of how we extract attention values and compute layerwise attention ratios are in Appendix~\ref{app:onevision}.

\begin{figure*}[!h]
    \centering
    \begin{minipage}[b]{0.48\linewidth}
        \centering
        \includegraphics[scale=0.25]{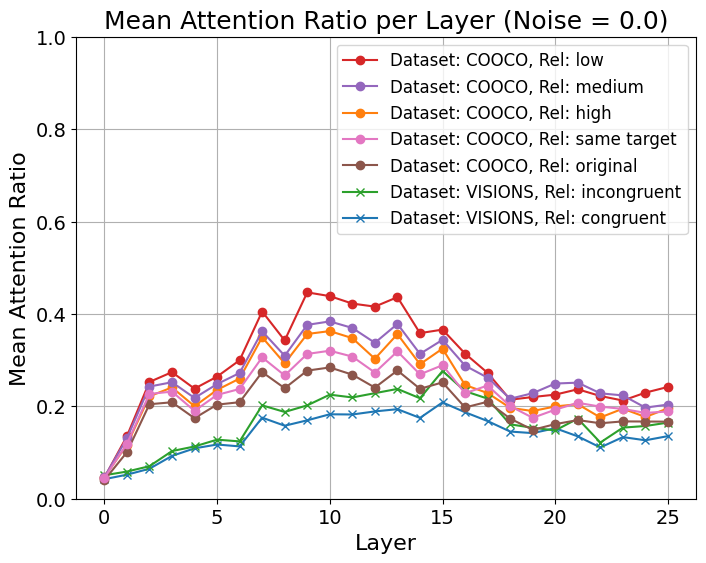}
        \caption*{\small \textbf{Correct Responses}}
    \end{minipage}
    \hfill
    \begin{minipage}[b]{0.48\linewidth}
        \centering
        \includegraphics[scale=0.25]{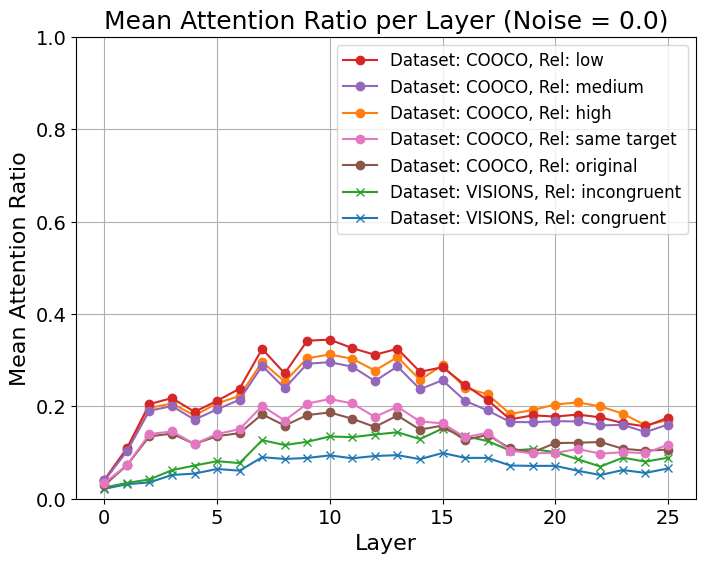}
        \caption*{\small \textbf{Incorrect Responses}}
    \end{minipage}
    \caption{Layer-wise attention allocation ratio on the target object across semantic relatedness conditions at zero noise, split by correct and incorrect responses.}
    \label{fig:attn_line}
\end{figure*}

\paragraph{Targets that violate scene semantics attract more attention.}
Figure~\ref{fig:attn_line} shows the attention allocation ratio across layers $\bar{r}_l$ for different semantic relatedness conditions at zero noise, separated by correct and incorrect responses according to the accuracy metric (\S~\ref{metrics}). We note several trends that hold in both COOCo and VISIONS datasets. 
First, we observe that, for correct (resp. incorrect) responses, the model generally applies more attention to the target. Second, target-focused attention is primarily concentrated in the middle layers, peaking between layers 7 and 15. Third, the model's attention to the target is clearly modulated by its semantic congruency with the scene: 
more attention is allocated in the \textit{low-relatedness} condition, followed by the \textit{medium-relatedness} condition, and least in the \textit{high-relatedness} condition. The least attention is allocated to the target in the \textit{original} condition, while the \textit{same target} condition receives slightly more. Interestingly, this pattern of increased attention to less semantically related targets also appears in incorrect responses. This suggests that even when the model fails to correctly name the object, it may still detect its incongruity with the surrounding scene, allocating more attention to targets that appear out of place. 
While the above observations apply to both datasets, we also note that $\bar{r}_l$ is generally lower in VISIONS. This is unlikely to be due to cues introduced by inpainting, as the difference also holds in comparison to the {\em original} image condition in COOCo. We hypothesise that the lowered attention-to-target in VISIONS is either an artefact of the image compression performed for our experiments, or is related to general properties of the images (e.g. the relative size of target regions). Deeper comparison of the differences between the two datasets is left to future work.

\begin{figure*}[!h]
    \centering
    \begin{minipage}[b]{0.48\linewidth}
        \centering
        \includegraphics[width=\linewidth]{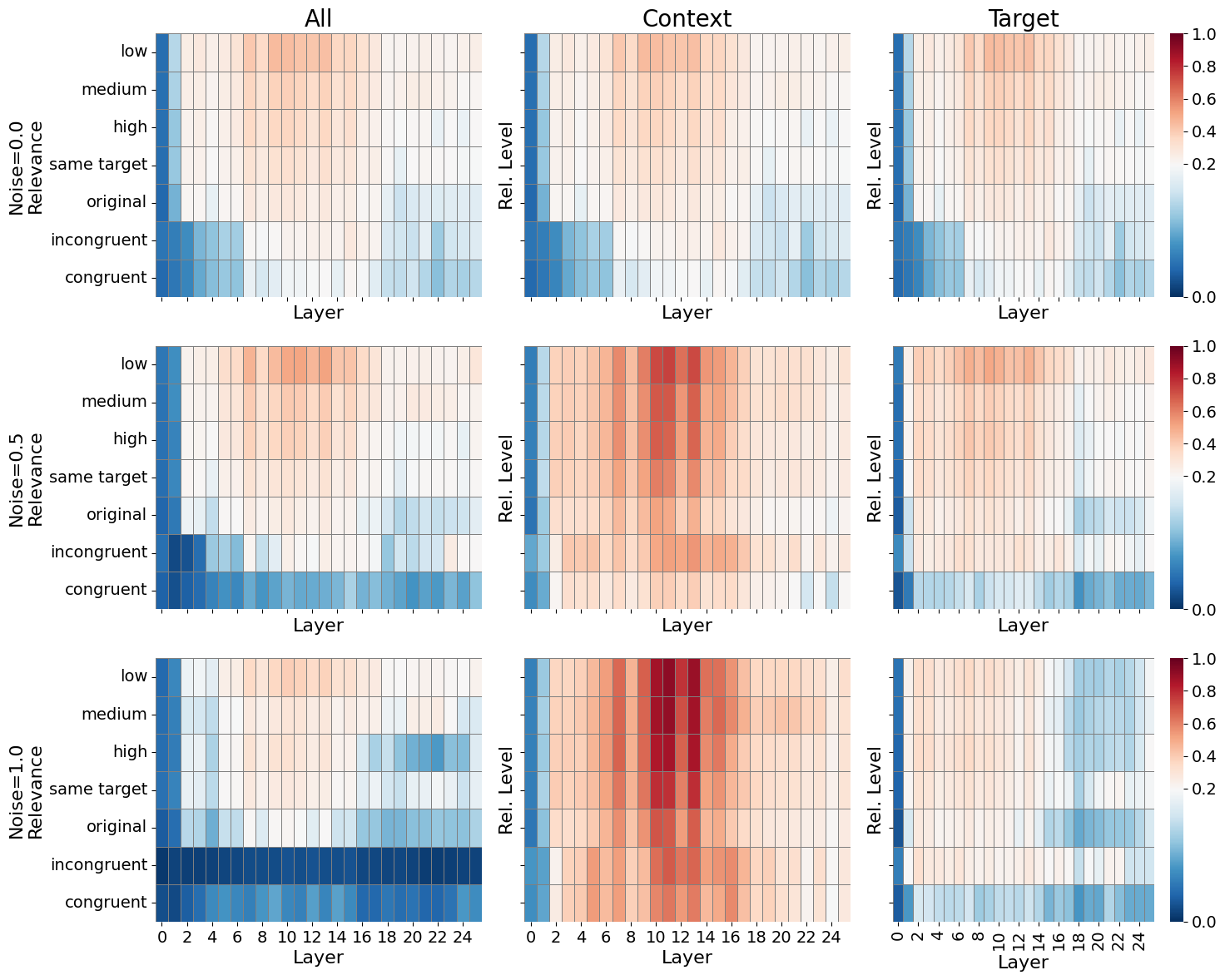}
        \caption*{\small \textbf{Correct Responses}}
    \end{minipage}
    \hfill
    \begin{minipage}[b]{0.48\linewidth}
        \centering
        \includegraphics[width=\linewidth]{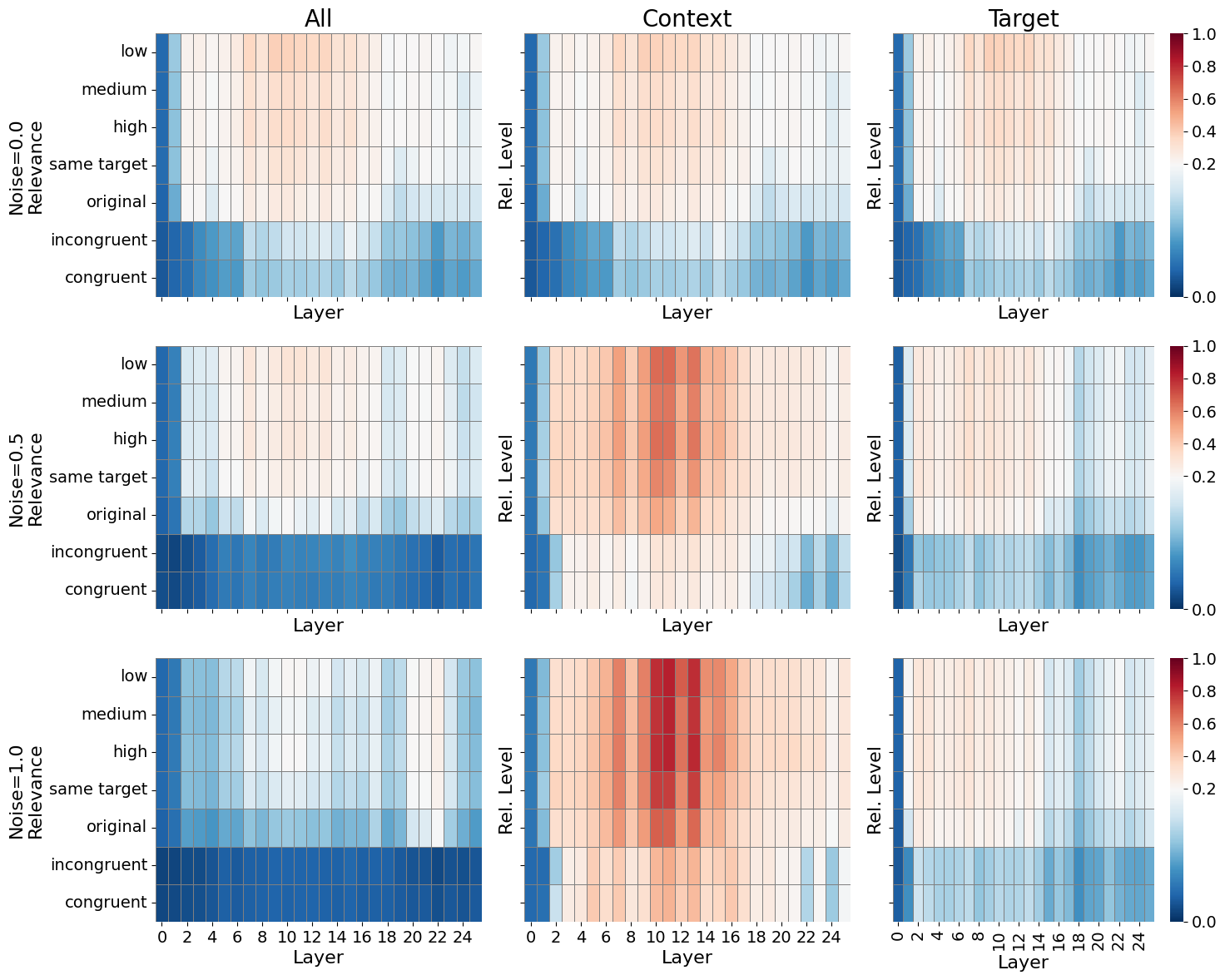}
        \caption*{\small \textbf{Incorrect Responses}}
    \end{minipage}
    \caption{Layer-wise attention heatmaps showing how the model's focus on the target varies with noise level (rows: none; moderate with $\lambda=0.5$; high with $\lambda=1.0$) and noise location (columns: all, context only, target only). Warmer colors indicate stronger attention relative to the no-noise baseline (average attention ratio $\sim$0.19), while cooler colors indicate reduced attention to the target. Target-scene relatedness levels range from {\em low} to {\em original} for COOCo, and {\em in/congruent} for VISIONS.}
    \label{fig:attn}
\end{figure*}

\paragraph{Targets attract attention even under moderate noise conditions.}
Figure~\ref{fig:attn} compares mean attention‐to‐target ratio $\bar{r}_l$ heatmaps for correct versus incorrect model responses across the three noise levels (0.0, 0.5, 1.0) and three noise conditions ({\em all, context, target}). 
The analysis shows that moderate noise on the target (e.g., noise level 0.5 in the {\em all}  and {\em target} noise conditions) tends to increase attention to the target in correct responses, particularly in the middle layers (7-14), suggesting that the model compensates for partial degradation by focusing more on the target. However, when the noise becomes too severe, attention to the target declines, likely because the target information becomes less recoverable. In contrast, under the {\em context} noise condition, higher noise levels consistently lead to increased attention on the target, indicating a reallocation of focus away from the noisy context toward the unaltered target. These attention patterns appear consistently in both COOCo and VISIONS. Once again we note a subtle difference between the two datasets: in the {\em all} noise condition, there is virtually no attention to the target region in the VISIONS samples. One possible reason, already alluded to above, is that inpainting in COOCo creates subtle cues that facilitate attention to the target region. However, in the COOCo original (non-inpainted) images we do observe some attention to the target, albeit much less so than in other conditions. These observations suggest that artefacts due to inpainting may facilitate attention to the target under {\em all} noise, but they are not the sole determining factor.
\newline
\noindent
Finally, we also see that when the model generates a correct response, there is higher attention to the target region, also under noisy conditions. Visualisations are in Appendix~\ref{app:results} (Fig.~\ref{fig:attn_delta_combined}). 

\begin{figure*}[!h]
    \centering
    \begin{minipage}[b]{0.48\linewidth}
        \centering
        \includegraphics[scale=0.25, trim=0 0 0 50, clip]{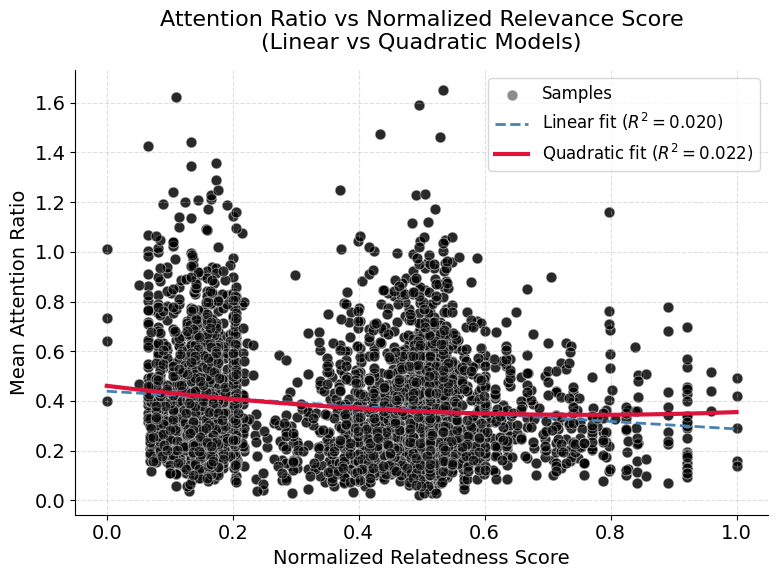}
        \caption*{\small \textbf{COOCo dataset}}
    \end{minipage}
    \hfill
    \begin{minipage}[b]{0.48\linewidth}
        \centering
        \includegraphics[scale=0.25, trim=0 0 0 50, clip]{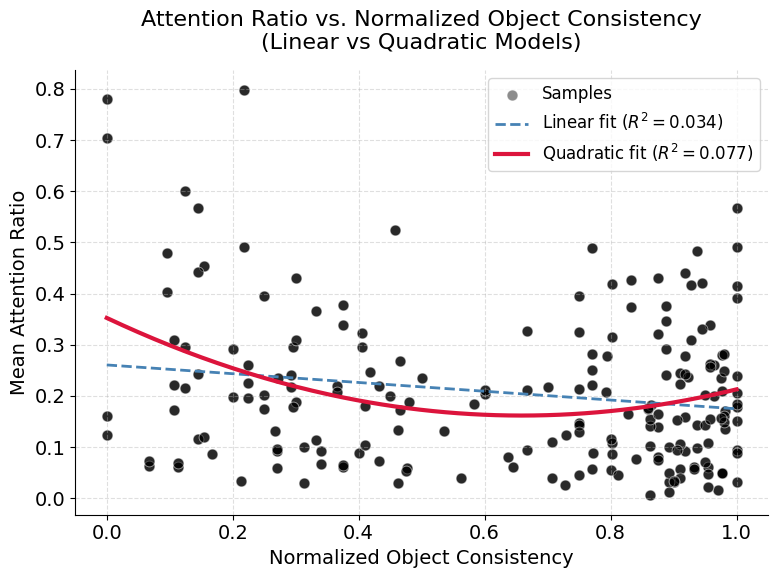}
        \caption*{\small \textbf{VISIONS dataset}}
    \end{minipage}
    \caption{Mean attention-to-target ratio as a function of semantic fit in COOCo (left) and VISIONS (right), aggregated over layers with consistently higher target focus. Points represent individual samples; dashed lines show linear regression fits, and solid red lines show quadratic fits.} 
    \label{fig:attn_vis}
\end{figure*}

\paragraph{Sensitivity to scene-target consistency is non-monotonic}
We further inspect how the attention-to-target ratio ($\bar{r}_l$) varies as a continuous function of scene–object semantic fit. For both COOCo and VISIONS, we restrict analyses to the samples where the model correctly identifies the target. In COOCo, we focus on manipulated images and exclude the \textit{original} and \textit{same-target} controls, as these do not involve modifications to the scene semantics; in VISIONS, we use the human-validated \emph{object-consistency ratings}.
We compute the mean attention-to-target ratio for each layer across the selected samples and identify the subset of layers that consistently show higher target focus (top quartile by mean attention). 
For each image, we then obtain a single attention score by averaging the attention-to-target ratio over this selected subset of layers. 
Semantic relatedness (COOCo) and object consistency (VISIONS) are normalised to the $[0,1]$ range and treated as continuous predictors. We model their relationship to the aggregated attention-to-target ratio using both linear and quadratic regression functions, and compare their goodness-of-fit via $R^2$. 
We present results for the 0 noise condition in Fig.~\ref{fig:attn_vis}; for results for the noised conditions refer to Appendix~\ref{app:results}, Fig.~\ref{fig:attn_continuous_all}.
At 0 noise, across datasets, attention toward the target exhibits a non-linear dependency on semantic fit. In VISIONS, a linear model revealed a significant negative association between object consistency and attention ($\beta = -0.0858$, $p < .001$), and adding a quadratic term significantly improved fit ($\beta_2 = 0.4395$, $p < .001$; nested model comparison: $F(1,597) = 27.75$, $p = 1.93\times10^{-7}$). The resulting curve was convex, with a minimum at $x = 0.659$ on the normalized scale, indicating reduced attention for moderately consistent objects and increased attention at higher consistency levels. A similar pattern emerged in COOCo. The linear model again showed a significant negative slope ($\beta = -0.1519$, $p < .001$), and the quadratic term provided a significant improvement ($\beta_2 = 0.2003$, $p < .001$; nested model comparison: $F(1,12051) = 19.86$, $p = 8\times10^{-6}$), with the minimum located at $x = 0.764$. While the effect is weaker in COOCo, both datasets exhibit reliable convex trends, suggesting that the semantic relatedness between target object and scene modulates attention in a non-monotonic manner rather than through a purely linear decline. This pattern is also present in other conditions for different noise levels and areas.

\section{Discussion}
\label{discussion}
The RefCLIPScore and Accuracy analyses reveal that models consistently exhibit superior performance when either the original target object or a semantically similar one is present. While it is possible that COCO images were included in models' training data, the observation that scores are slightly higher for the \emph{same target} condition than for the \emph{original} condition suggests that models benefit from small semantic shifts within the same category, potentially because the generated object is more prototypical and thus easier to categorise. Alternatively, inpainting introduces artefacts that may make the target object more salient, thereby drawing additional model attention. This could be mitigated in future work by using more advanced, frontier inpainting models. However, the decrease in performance with reduced object--scene relatedness indicates that models are sensitive to target-context semantic compatibility.
Noise manipulations provide additional insight: When targets are weakly related to the scene, reducing contextual information improves performance, whereas 
for semantically congruent targets, preserving scene context is more beneficial when the target region is degraded.

The output versus scene similarity results for LLaVA-OneVision indicate that as semantic relatedness decreases, the model tends to anchor outputs more to the semantics of the target rather than the broader scene. When target objects are corrupted by noise, correct responses shift towards a stronger reliance on scene context, implying an adaptive strategy that leverages global context to compensate for degraded local features, consistent with \citet{junker-zarriess-2024-resilience-scene}.

In the attention analysis, clear differences related to semantic fit emerge in the Transformer decoder’s attention allocation. Correct categorisation consistently involves higher attention to the target, particularly in middle layers (7–15), suggesting that critical semantic processing occurs there, as already noted in interpretability studies on the LLaVA model family \cite{zhang_mllms_2025, zhang_cross-modal_2025, golovanevsky_what_2025}. A compelling observation is the inverse relationship between semantic relatedness and attention allocation: less scene-related targets receive more focused attention. Furthermore, noise conditions modulate attention dynamics significantly. Moderate target-specific noise (0.5 level) triggers a compensatory increase in target-focused attention, aiding correct recognition. Conversely, extreme noise (level 1.0) forces models to shift reliance entirely to context. This dynamic attention behaviour highlights an adaptive mechanism within the model, balancing local target analysis against global context integration depending on image clarity and semantic coherence. Finally, LLaVA-One-Vision’s non-monotonic sensitivity to scene–object fit has a notable parallel in work by \citet{damiano_exploring_2024}, whose eye-tracking study shows that both highly expected and unexpected objects receive increased fixations~\cite[see also][]{celikkol_tf-idf_2023}. The resulting U-shaped relationship between semantic similarity and overt attention has an analogous trend in our results: moderately consistent targets attract reduced attention, whereas both low- and high-consistency targets receive higher focus. While the underlying mechanisms clearly differ, this similarity indicates that the model’s attention patterns resemble key aspects of human scene processing.

\section{Conclusion}
\label{conc}
This paper presented COOCo, a dataset to explore scene-semantic violations.
Consistent with recent work~\cite{junker-zarriess-2024-resilience-scene,junker_scenegram_2025}, we show that VLMs rely on scene semantics to refer to objects, but this is modulated by perturbations to these image regions. While these results hold across different models, and are supported by in-depth analysis of LLaVA-OneVision-0.5B, we believe that detailed comparison of different architectures would shed further light on the role of context in VLMs. We replicated our findings on VISIONS~\cite{allegretti_visual_2025}, which differs from COOCo in that it does not involve artificial image manipulation.

Our analysis of attention under different conditions contributes to a growing body of interpretability research aimed to localise cross-modal fusion in VLMs~\cite{zhang_mllms_2025, zhang_cross-modal_2025, golovanevsky_what_2025}.

\section*{Acknowledgements}
This research was supported by the Netherlands Science Foundation (NWO) under the AiNed XS Europe 2023 program, grant number NGF.1609.23.020. FM was a visiting student from the University of Trento to Utrecht University during the period when the work was conducted. We thank Sina Zarrieß, Simeon Junker and Hendrik Buschmeier for their valuable comments and helpful discussions. We also thank our four anonymous TACL reviewers and the action editor, whose input helped to improve the paper.

\bibliography{custom}
\bibliographystyle{acl_natbib}

\clearpage

\appendix
\section{Additional preprocessing, data and model details}

\subsection{COOCo image creation}\label{app:scenes}
\begin{table}[!h]
    \centering
    \begin{tabularx}{\columnwidth}{|p{7.25cm}|}
    \hline 
         living room, arrival gate outdoor, restaurant, market/outdoor, parking lot, bakery shop, arrival gate/outdoor, art studio, gas station, banquet hall, tower, street, music studio, fastfood restaurant, pantry, cubicle/office, market outdoor, kitchen, coffee shop, bedroom, dorm room, cubicle office, delicatessen, train railway, bakery/shop, home office, physics laboratory, dining room, bathroom.\\
    \hline
    \end{tabularx}
    \caption{The 25 scene labels in the COOCo dataset}
    \label{tab:scene-labels}
\end{table}

\begin{figure}[h!]
    \centering
    \includegraphics[width=\linewidth, trim=0 0 0 35, clip]{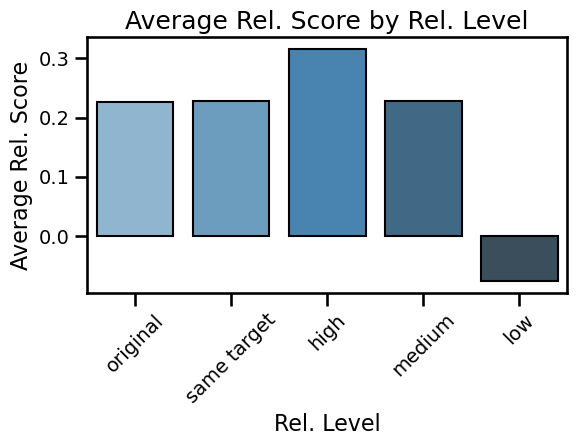}
    \caption{Average relatedness scores grouped by relatedness level, illustrating the semantic similarity between scenes and candidate objects. 
    }
    \label{fig:scene_obj_rel}
\end{figure}

\subsubsection{Inpainting and verification pipeline}
We use the LaMa inpainting model \cite{suvorov_resolution-robust_2021} to remove the target object from a scene, expanding its mask by 20\% to make sure that the resulting image is free of irregularities. We then create a square-padded version of the cleaned image, and create a mask for object inpainting, 
For each of the three relatedness groups (low, medium, and high), we retain 15 replacement candidates. In the high-similarity condition only, the similarity to the original target object is also considered.
Additionally, we generate one image containing the same target object as in the original image to use as a control condition. See Figure~\ref{fig:scene_obj_rel} for the distributon of relatedness scores per relatedness level.

We randomly sample one candidate, and generate a descriptive prompt using the 8-bit quantised \cite{jacob_quantization_2017} LLaVA model \cite{liu_visual_2023}. We perform inpainting using PowerPaint \cite{zhuang_task_2024}, 
guiding the inpainting process with the LLaVA-generated descriptive prompt as a positive prompt and a fixed negative prompt to suppress undesired content. We set the fitting degree (which controls how closely the output adheres to the shape of the mask) to 0.6 (0-1), use 45 (0-50) diffusion steps to balance quality and compute efficiency, and apply a guidance scale of 7.5 (0-30) that controls for alignment with the prompt.

We verify resulting images using LLaVA through a binary visual query assessing the presence of the generated candidate object in the image; if rejected, the guidance scale is increased by 7.5, and the process is repeated up to four times until the maximum guidance scale level of 30. If all attempts fail, a new candidate is sampled. This process yields 8 generated images per input (3 each for the low- and medium relatedness conditions, 1 for high relatedness, and 1 for the same-target condition). A second round of filtering is applied using full-precision LLaVA to ensure final quality. 

\subsection{Spatial encoding \& model prompts}\label{app:preproc}

\begin{tcolorbox}[colback=gray!15!white, colframe=black, title=KOSMOS-2]
KOSMOS-2 encodes spatial regions by discretizing bounding boxes into location tokens and embedding them alongside natural language using \texttt{<grounding>} tags.

\textbf{Prompt:}
\begin{quote}
<grounding>What is the object in <phrase>this part</phrase> of the image? Answer with the object's name only. No extra text. Answer:
\end{quote}
\end{tcolorbox}

\paragraph{KOSMOS-2} KOSMOS-2 employs a discretized token-based representation of spatial information. The key principle is to transform continuous image coordinates into discrete location tokens via a grid-based partitioning of the image space. Each image is divided into a $P \times P$ grid, and every bounding box is encoded by mapping its top-left and bottom-right corners to location tokens. These tokens are then embedded in the textual input using a structured format that mimics hyperlink annotations, explicitly marking visual references within language. This design enables KOSMOS-2 to align and jointly reason over linguistic spans and visual regions in a shared sequence modeling framework.

\paragraph{Molmo}
Molmo grounds language in visual space using a point-based strategy. Instead of bounding boxes, it represents spatial references as single $(x, y)$ coordinates, normalized to a fixed-scale [0,100] grid. Such grounding enables the model to identify, count, or compare objects based on minimal visual supervision. The model interprets these point references directly from the linguistic input, treating them as spatial queries in the generation process.

\begin{tcolorbox}[colback=gray!15!white, colframe=black, title=Molmo]
Molmo grounds text in visual space using single-point references normalized to a [0,100] scale. These coordinates are embedded directly in the natural language prompt.

\textbf{Prompt:}
\begin{quote}
What is the object at point x = 63, y = 44 of the image? Answer with the object's name only. No extra text.
\end{quote}
\end{tcolorbox}

\paragraph{xGen-MM (BLIP-3)}

\noindent
xGen-MM encodes spatial grounding using natural language expressions that describe object locations via bounding box coordinates. 

\begin{tcolorbox}[colback=gray!15!white, colframe=black, title=xGen-MM]
xGen-MM embeds bounding box coordinates within the prompt using \texttt{<bbox>} tags and learns to interpret these as visual region references in natural language form.

\textbf{Prompt:}
\begin{quote}
<|user|>  
<image>What is the object in this part <bbox>[x1, y1][x2, y2]</bbox> of the image? Answer with the object's name only. No extra text.<|end|>  
<|assistant|>
\end{quote}
\end{tcolorbox}

\paragraph{Qwen2.5-VL}
Qwen2.5-VL builds upon the Qwen-VL architecture by integrating visual tokens into a large language model using enhanced spatial encoding strategies. It employs bounding box references enclosed in square bracket format $[x_1, y_1, x_2, y_2]$, where coordinates are normalized within the range [0,1]. The spatial references are directly inserted into the natural language prompt, making them interpretable as explicit multimodal instructions.

\begin{tcolorbox}[colback=gray!15!white, colframe=black, title=Qwen2.5-VL]
Qwen2.5-VL encodes normalized bounding box coordinates directly in the prompt as $[x_1, y_1, x_2, y_2]$, supporting explicit multimodal alignment through instruction-tuned sequences.

\textbf{Prompt:}
\begin{quote}
What is the object in this part of the image [x1, y1, x2, y2]? Answer with the object's name only. No extra text.
\end{quote}
\end{tcolorbox}

\paragraph{LLaVA-OneVision}
LLaVA-OneVision employs a normalized bounding box encoding strategy embedded directly in the textual prompt in the form \texttt{[x1, y1, x2, y2]}, where all values fall within the [0,1] range.

\begin{tcolorbox}[colback=gray!15!white, colframe=black, title=LLaVA-OneVision]
LLaVA-OneVision embeds bounding boxes in [0,1] range directly in the prompt, leveraging instruction-tuned pretraining to interpret spatial references as tasks.

\textbf{Prompt:}
\begin{quote}
What is the object in this part of the image [x1, y1, x2, y2]? Answer with the object's name only. No extra text.
\end{quote}
\end{tcolorbox}

\subsection{Accuracy comparison between dataset labels and modal object names in VISIONS}
\label{app:modal_names}

\begin{table}[h!]
\centering
\caption{Contingency table showing the counts and row percentages of agreement and disagreement between accuracy computed with dataset labels (rows) and with modal names (columns), along with the overall correlation and number of divergent cases.}
\label{tab:accuracy_contingency}

\subsection*{Counts}
\begin{tabular}{lrr}
\hline
\textbf{Accuracy} & \textbf{0} & \textbf{1} \\
\hline
\\
\quad\quad\quad\quad\quad\quad\quad\quad\quad\quad\quad \textbf{0} & 4817 & 73 \\
\textbf{Accuracy Modal Names}\\
\quad\quad\quad\quad\quad\quad\quad\quad\quad\quad\quad \textbf{1} & 95   & 955 \\
\hline
\end{tabular}

\vspace{0.5em}
\subsection*{Row Percentages}
\begin{tabular}{lrr}
\hline
\textbf{Accuracy} & \textbf{0} & \textbf{1} \\
\hline
\\
\quad\quad\quad\quad\quad\quad\quad\quad\quad\quad\quad \textbf{0} & 98.1 & 7.1 \\ 
\textbf{Accuracy Modal Names}\\
\quad\quad\quad\quad\quad\quad\quad\quad\quad\quad\quad \textbf{1} &  1.9 & 92.9 \\
\hline
\end{tabular}

\vspace{0.5em}
\begin{flushleft}
\textbf{Pearson correlation:} 0.902 \\
\textbf{Divergent cases:} 168 / 5940 (2.83\%)
\end{flushleft}

\end{table}

We use the modal names provided in the VISIONS dataset to compare our cosine-similarity-based accuracy metric under two different ground truths: (1) the dataset labels and (2) the human-generated modal names. This comparison allows us to assess how sensitive the metric is to variation in naming.
The contingency tables \ref{tab:accuracy_contingency} show a strong alignment between the two ground-truth conditions. Predictions judged incorrect using the dataset label are almost always incorrect using the modal name as well (98.1\%), and predictions judged correct with the dataset label are likewise correct with the modal name in most cases (92.9\%). The low divergence rate (2.83\%) and high correlation (0.902) indicate that disagreements between the two ground-truth settings are rare and that the metric captures nearly identical performance patterns across them. Overall, these results suggest that the accuracy metric is robust to synonym variation and is not substantially affected by differences in naming between dataset labels and human-provided modal names.

\subsection{Attention over image tokens in LLaVA-OneVision}\label{app:onevision}
LLaVA-OneVision handles an image–question pair in four stages: 

\begin{enumerate}
    \item Following the AnyRes visual tokenization strategy, the input image is resized and tiled into $a\times b$ crops of the same resolution. Each view -- including the full image -- is encoded by the SigLIP SO400M ViT \cite{zhai_sigmoid_2023} into $T$ tokens, yielding $L = (a\times b + 1)\,T$ tokens in total. If $L$ surpasses the threshold $\tau$, bilinear interpolation reduces the per-crop token count.
    \item All visual tokens are projected via a two‐layer MLP into the LLM’s embedding space;
    \item The resulting image tokens are prepended to the tokenized question together with an answer‐start marker;
    \item The LLM then decodes the answer autoregressively.
\end{enumerate}

In Section~\ref{exp_attention}, we restrict our analysis to the visual tokens obtained from the fixed encoding of the full image, omitting any multi‐crop tokens.

Below, we detail the process of aggregating attention values and computing the layerwise attention allocation ratio $r^{(t)}_{l}$ described in Section~\ref{exp_attention}.

\subsection{Aggregating LLM Attention}
We first extract the attention deployed by the LLM decoder over image tokens.
Let the attention tensor be $A \in \mathbb{R}^{L \times H \times N \times N}$, where $L$ is the number of layers, $H$ is the number of attention heads, and $N$ is the number of tokens. For each layer $l$, we compute the mean attention over heads: $\bar{A}_l = \frac{1}{H} \sum_{h=1}^H A_{l,h}$.
To improve interpretability, we follow a common practice of zeroing out attention directed toward the beginning-of-sequence (BOS) token, which often absorbs a disproportionate amount of attention due to its special role in autoregressive models~\cite{vig_analyzing_2019}. Specifically, for each attention matrix $\bar{A}_l$, we set the column corresponding to the BOS token to zero, effectively removing all incoming attention to this token. We then normalize each row so that its values sum to one, resulting in a normalized attention matrix $\hat{A}_l \in \mathbb{R}^{N \times N}$.

We aggregate attention across layers to obtain the final LLM attention matrix: $A_{\mathrm{LLM}} = \frac{1}{L} \sum_{l=1}^L \hat{A}_l$. Our method distinguishes between prompt tokens and generated tokens in accordance with the autoregressive nature of the model. Prompt tokens receive attention from all subsequent tokens, while generated tokens attend only to preceding tokens. To construct the attention representation, we first process the prompt tokens by computing their attention matrix as described above.

For generated tokens, we apply a similar procedure with two key modifications: (1) only the last row of each attention matrix is retained, corresponding to the attention from the newly generated token, and (2) attention to the BOS token is nullified as before. Each resulting attention vector is then padded to a common length and stacked with the aggregated prompt token attention matrix.

\subsection{Aggregating Attention in Vision Transformers (ViTs)}
During our initial experiments, we noticed a tendency where the model applied attention over a specific set of tokens for many images, indicating a model-intrinsic bias. To mitigate such systematic attention biases, we compute a layer-wise average attention map across the dataset and subtract it from each instance’s attention map. This adjustment emphasizes instance-specific patterns by removing model-intrinsic tendencies. To avoid resolution-based confounds, we include only images of size 640$\times$640 (which comprise 86\% of the dataset).

\textbf{Per-Instance Layer Aggregation}:  
For each image instance $t \in T$ and layer $l \in L$, we average the attention across heads: $\bar{A}_l^{(t)} = \frac{1}{H} \sum_{h=1}^H A_{l,h}^{(t)}$. We then normalize each row of $\bar{A}_l^{(t)}$ so that attention values sum to one, yielding a normalized attention map $A_{\mathrm{norm}, l}^{(t)}$.

\textbf{Dataset-Wide Averaging}:  
To capture general attention patterns, we compute the average normalized attention map across all instances: $A_{\mathrm{avg}, l} = \frac{1}{T} \sum_{t=1}^T A_{\mathrm{norm}, l}^{(t)}$.
These maps serve as baselines for interpreting the attention behavior of individual instances. The final per-instance attention map is computed by subtracting the baseline and applying ReLU: $A_{\mathrm{ViT}, l}^{(t)} = \mathrm{ReLU} \left( A_{\mathrm{norm}, l}^{(t)} - A_{\mathrm{avg}, l} \right)$.
This operation retains only positive deviations from the dataset-wide mean, highlighting unique attention patterns for each instance. Each column of $A_{\mathrm{ViT}, l}^{(t)}$, corresponding to a vision token $j \in J$, is reshaped into a spatial attention map $V_j \in \mathbb{R}^{g \times g}$, where $g$ is the number of image patches per side (e.g., $g = 27$ for SO400M). These maps represent spatial distributions of attention at the selected layer.

\subsection{Attention Over Image Regions}

To analyze how the model distributes its focus across visual inputs during generation, we compute attention over image regions.

Let $O = \{o_1, o_2, \ldots, o_{|O|-1}\}$ denote the set of the LLM decoder output tokens, excluding the final \texttt{<|im\_end|>} token, which is not semantically informative. For each output token $o_i \in O$, we extract the corresponding attention weights $A_{\mathrm{LLM}, i, j}$ from the LLM attention matrix, where $j$ indexes the vision tokens. These attention weights are normalised to yield $\hat{A}_{i,j}$.

Each vision token $j$ is associated with a spatial map $V_j \in \mathbb{R}^{g \times g}$, derived from the adjusted ViT attention described earlier. The attention map over the image for output token $o_i$ is then computed as a weighted sum: $A_i^{\mathrm{img}} = \sum_{j \in J} \hat{A}_{i,j} \cdot V_j$.
To obtain a global attention map for the image, we average over all output tokens: $A_{\mathrm{final}}^{\mathrm{img}} = \frac{1}{|O| - 1} \sum_{i \in O} A_i^{\mathrm{img}}$.
This attention map is then upsampled to the original image resolution using nearest-neighbor interpolation for visualization.

\subsection{Attention Over Target and Context}

To analyze how attention is distributed between the target object and the surrounding context, we extract the bounding box $B = (x_{\min}, y_{\min}, w, h)$ of the target object region. The total attention over the target area is computed as: $a_{\mathrm{target}} = \sum_{(x,y) \in B} A_{\mathrm{final}}^{\mathrm{img}}(x,y)$.
Similarly, the attention over the context region, excluding both the target and irrelevant padding areas, is given by: $a_{\mathrm{context}} = \sum_{(x,y) \in C} A_{\mathrm{final}}^{\mathrm{img}}(x,y)$,
where $C$ represents the pixels outside the target bounding box but within the valid image region.

The entire process is carried out for each layer of the vision backbone model, for each of which we have a specific aggregated attention representation $A_{\mathrm{ViT}, l}$, ensuring that we obtain an attention distribution over the image, and thus over both the target and context, specific to each layer. 

\subsection{Attention Allocation Ratio}  
To analyze attention allocation across different levels \( l \) of the vision encoder and experimental instances \( t \), we compute the ratio \( r \) between the total normalized attention values assigned to the target (\( a_{\mathrm{target}} \)) and the context (\( a_{\mathrm{context}} \)): $r_l^{(t)} = \frac{a_{\mathrm{target},l}^{(t)}}{a_{\mathrm{context},l}^{(t)}}$.
A value of \( r \) closer to 1 indicates a higher allocation of attention to the target region; a value closer to 0 suggests greater attention to the context. Since the context area is larger than the target, we do not expect \( r \) to exceed 0.5. Instead, our focus is on analyzing its variations across different experimental conditions.

We then compute the mean ratio value for each encoder layer by averaging across all instances: $\bar{r}_l = \frac{1}{T} \sum_{t=1}^{T} r_l^{(t)}$
where \( T \) denotes the total number of instances. This allows us to assess how attention is distributed across different hierarchical levels of the encoder under varying experimental conditions.

\section{Complementary results}\label{app:results}

\begin{table}[!h]
\centering
\tiny
\begin{tabular}{l|ccc|ccc}
\toprule
    & \multicolumn{3}{c|}{\textbf{Congruent images}} & \multicolumn{3}{c}{\textbf{Incongruent images}}\\
 & \textbf{Coef.} & \textbf{Std.Err.} & \textbf{z} & \textbf{Coef.} & \textbf{Std.Err.} & \textbf{z} \\
\midrule
Int. & 1.744 & 0.054 & 32.406$^*$ & 1.745 & 0.055 & 31.902$^*$ \\
Context & -0.087 & 0.013 & -6.56$^*$ & -0.178 & 0.013 & -13.518$^*$ \\
Target & -0.105 & 0.013 & -7.866$^*$ & -0.094 & 0.013 & -7.152$^*$ \\
N. level & -0.040 & 0.009 & -4.261$^*$ & -0.028 & 0.009 & -2.953$^*$ \\
N. level $\times$ Con & 0.009 & 0.013 & 0.667 & -0.001 & 0.013 & 0.044 \\
N. level $\times$ Tar & 0.041 & 0.013 & 3.030$^*$ & 0.023 & 0.013 & 1.78 \\
Group Var& 0.020 & 0.044 &  & 0.020 & 0.046 &  \\
\bottomrule
\end{tabular}
\caption{\footnotesize Linear mixed effects model on VISIONS: Noise area (Context/Target) and noise level (N. level) on scene-target semantic similarity, for congruent and incongruent conditions. $^*$ denotes a significant effect at $p \leq .01$. Conditional $R^2$ values: 0.423 (congruent) and 0.452 (incongruent).}
\label{tab:lme-visions}
\end{table}

\begin{figure}[!h]
    \centering
    \includegraphics[scale=0.25, trim=0 0 0 0, clip]{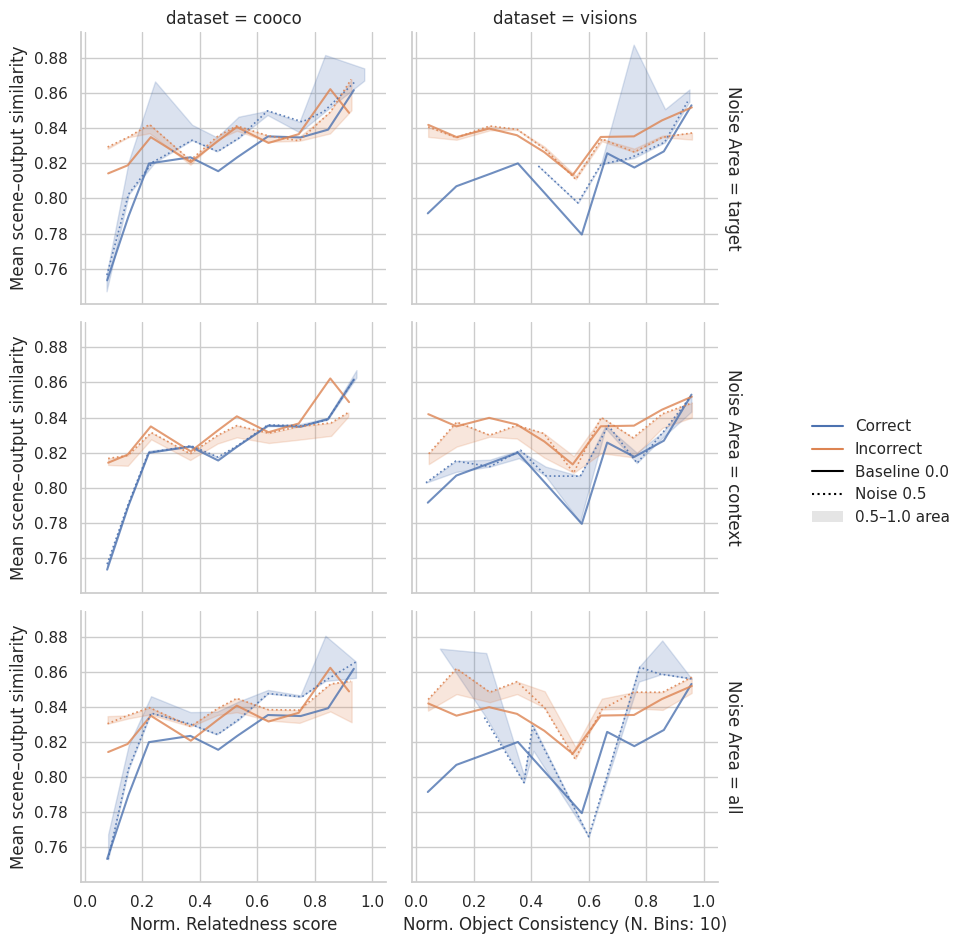}
    \caption{\footnotesize Mean semantic similarity between model outputs and scene labels as a function of binned continuous object–scene relatedness (COOCo) and object consistency (VISIONS), using 10 bins. Columns correspond to datasets; rows to noise–area conditions (target, context, all). Continuous lines represent the baseline (0.0 noise) for correct (blue) and incorrect responses (orange); dotted lines indicate the 0.5-noise condition; shaded regions show the variability introduced when increasing noise from 0.5 to 1.0.}
    \label{fig:scene_vs_output_continuous}
\end{figure}

\begin{figure}[!t]
    \centering
    \begin{subfigure}{\linewidth}
        \centering
        \includegraphics[scale=0.13]{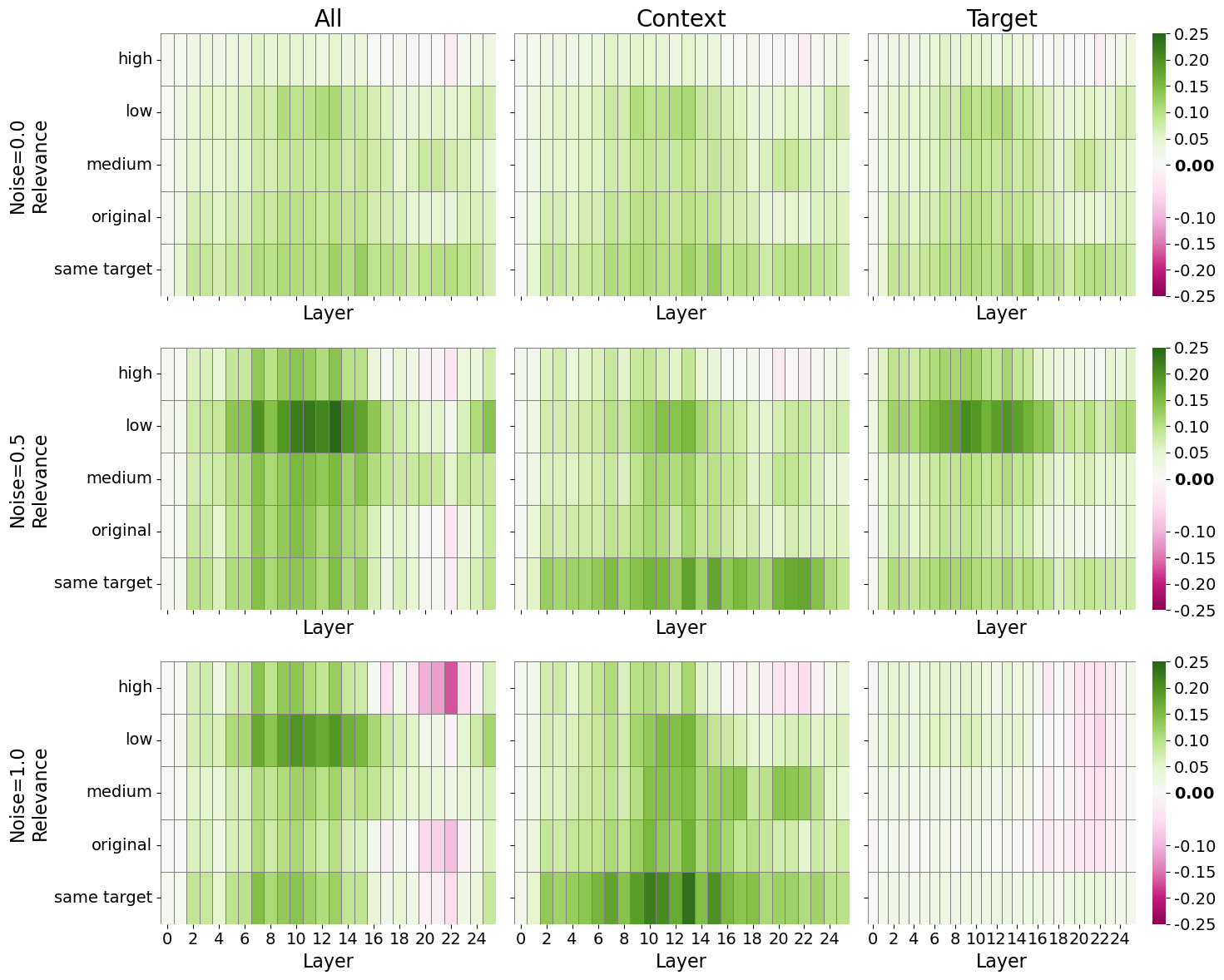}
        \caption{COOCo, $\Delta \bar{r}_l$ (range: $\pm 0.25$)}
        \label{fig:attn_delta_cooco}
    \end{subfigure}
    
    \begin{subfigure}{\linewidth}
        \centering
        \includegraphics[scale=0.18]{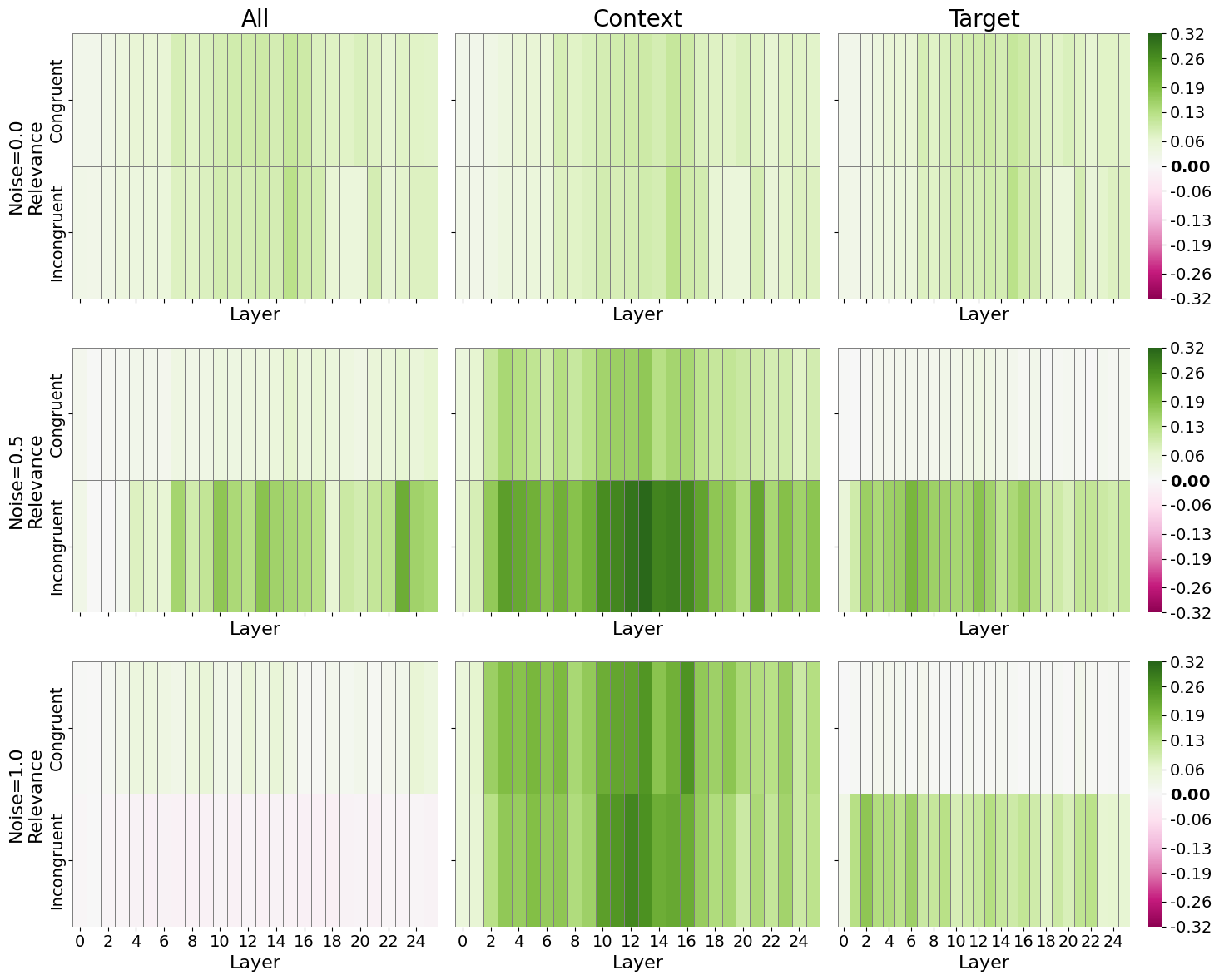}
        \caption{VISIONS, $\Delta \bar{r}_l$
        (range: $\pm 0.32$).}
        \label{fig:attn_delta_visions}
    \end{subfigure}
    \caption{\footnotesize Attention-to-target delta $\Delta \bar{r}_l$, between correct and incorrect predictions across layers, semantic relatedness levels, noise levels, and noise areas. Heatmaps are predominantly green, indicating that when the model correctly recognizes the object, it tends to allocate more attention to the target token compared to when it misclassifies the object. 
    }
    \label{fig:attn_delta_combined}
\end{figure}

\begin{figure*}[!h]
    \centering

    \textbf{COOCo dataset}\\[6pt]

    \begin{minipage}[b]{0.32\linewidth}
        \centering
        \includegraphics[width=\linewidth, trim=0 0 0 50, clip]{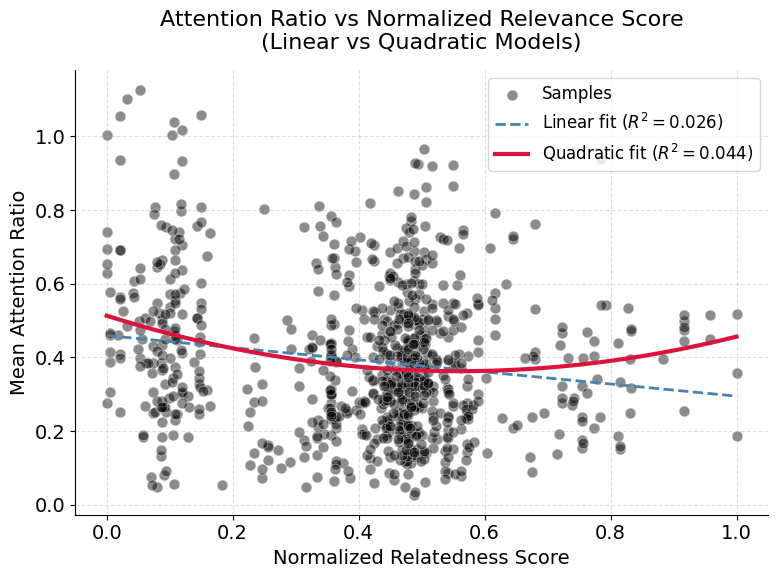}
        \caption*{\small 0.5 noise – Target}
    \end{minipage}
    \begin{minipage}[b]{0.32\linewidth}
        \centering
        \includegraphics[width=\linewidth, trim=0 0 0 50, clip]{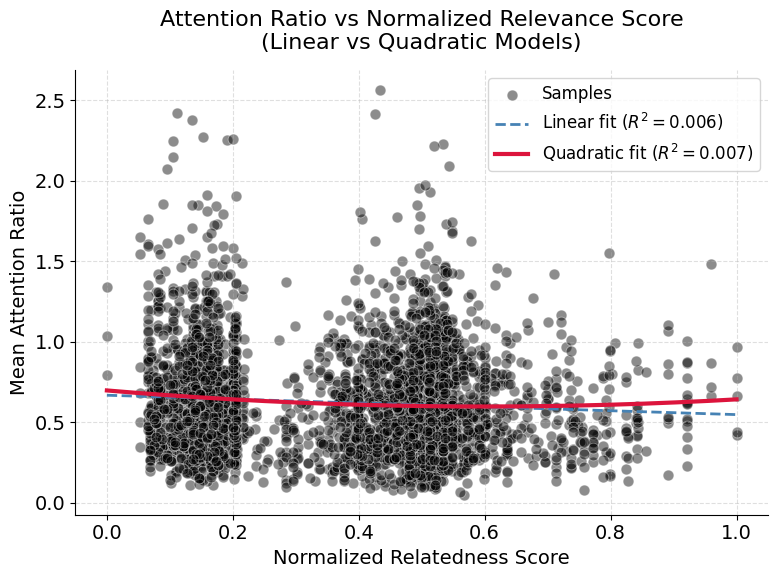}
        \caption*{\small 0.5 noise – Context}
    \end{minipage}
    \begin{minipage}[b]{0.32\linewidth}
        \centering
        \includegraphics[width=\linewidth, trim=0 0 0 50, clip]{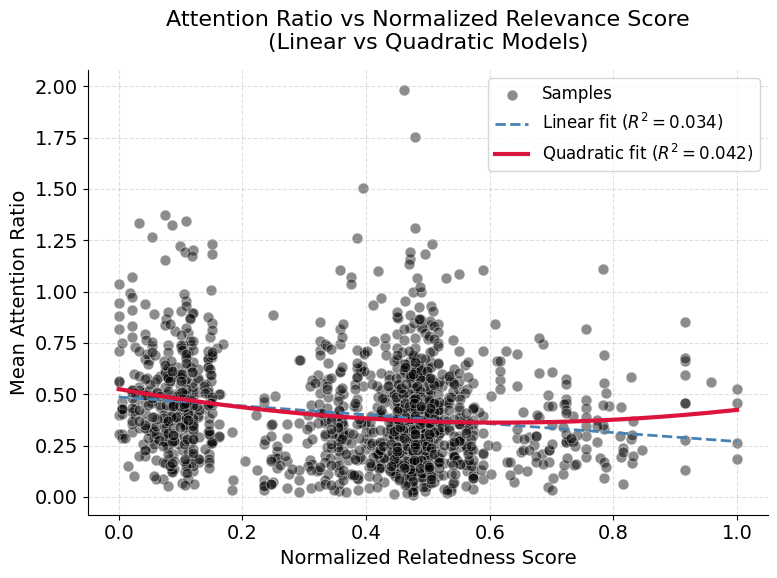}
        \caption*{\small 0.5 noise – All}
    \end{minipage}

    \vspace{10pt}

    \begin{minipage}[b]{0.32\linewidth}
        \centering
        \includegraphics[width=\linewidth, trim=0 0 0 50, clip]{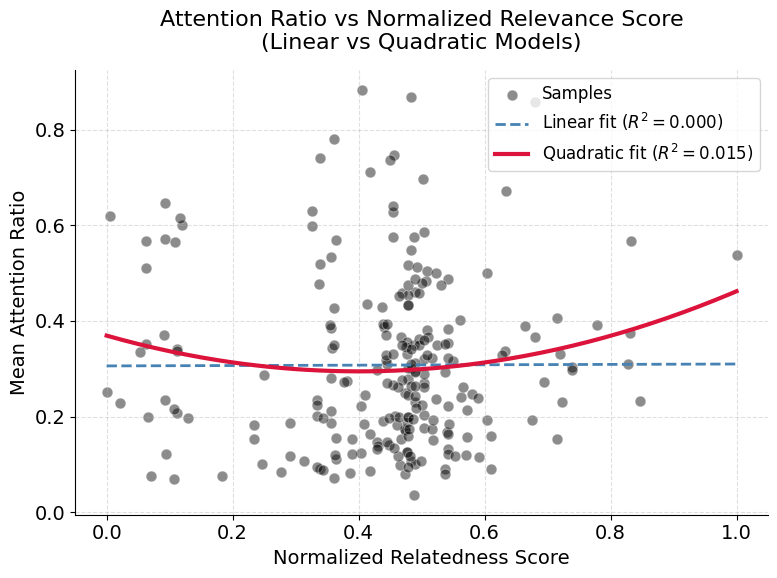}
        \caption*{\small 1 noise – Target}
    \end{minipage}
    \begin{minipage}[b]{0.32\linewidth}
        \centering
        \includegraphics[width=\linewidth, trim=0 0 0 50, clip]{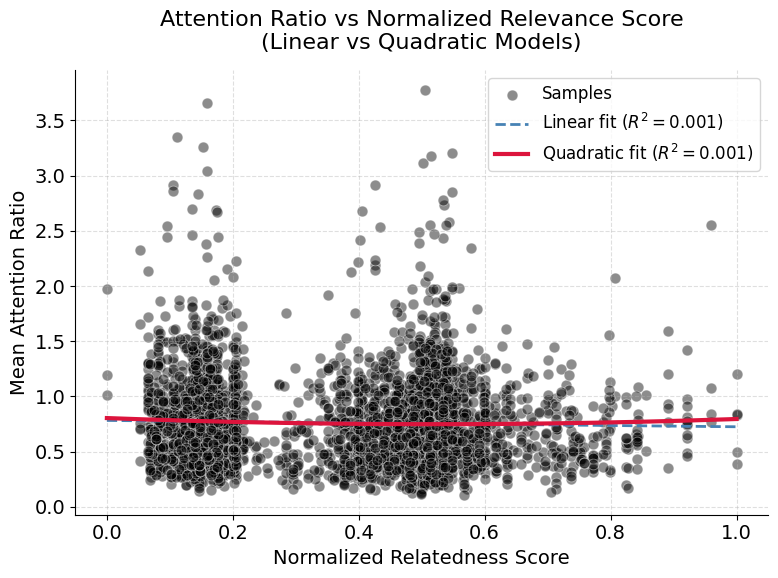}
        \caption*{\small 1 noise – Context}
    \end{minipage}
    \begin{minipage}[b]{0.32\linewidth}
        \centering
        \includegraphics[width=\linewidth, trim=0 0 0 50, clip]{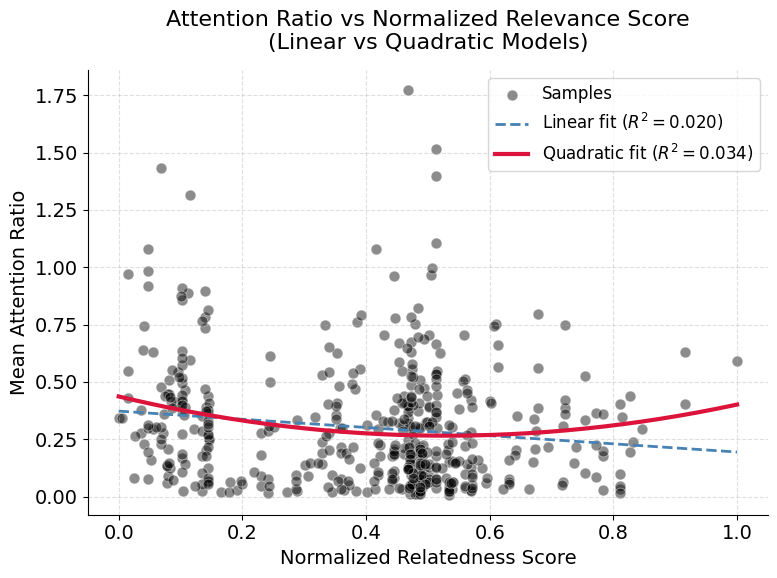}
        \caption*{\small 1 noise – All}
    \end{minipage}

    \vspace{14pt}

    \textbf{VISIONS dataset}\\[6pt]

    \begin{minipage}[b]{0.32\linewidth}
        \centering
        \includegraphics[width=\linewidth, trim=0 0 0 50, clip]{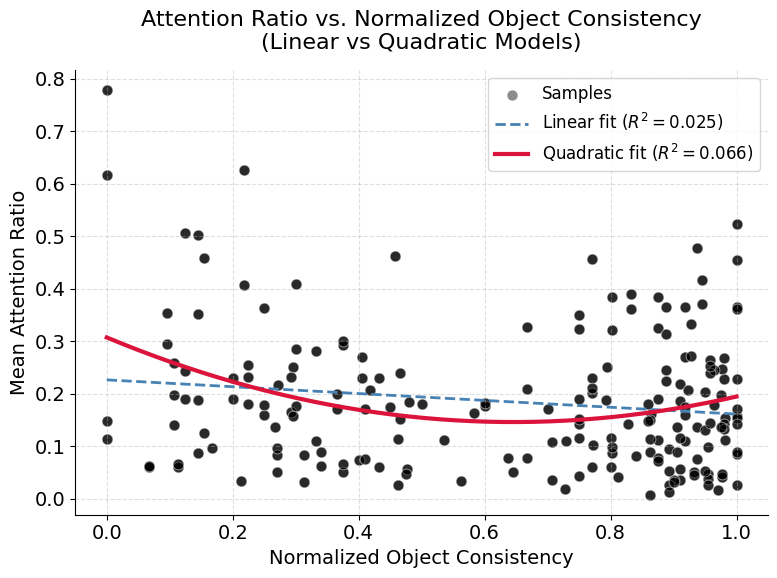}
        \caption*{\small 0.5 noise – Target}
    \end{minipage}
    \begin{minipage}[b]{0.32\linewidth}
        \centering
        \includegraphics[width=\linewidth, trim=0 0 0 50, clip]{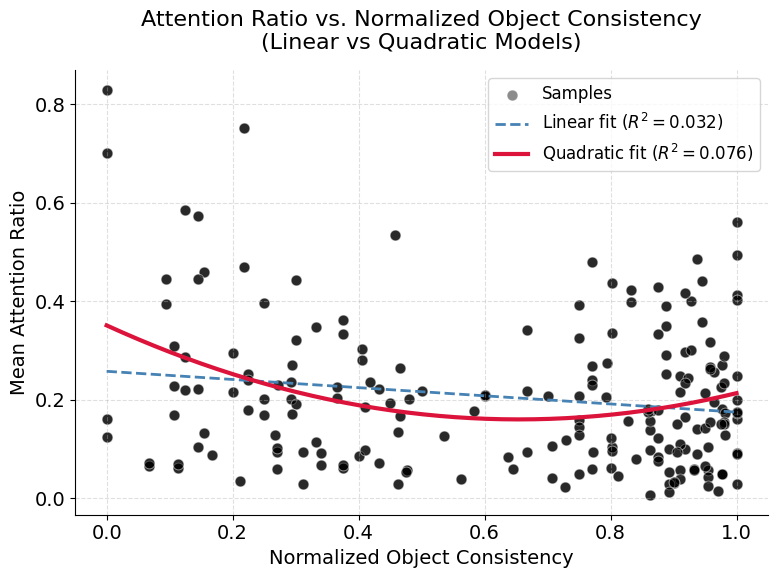}
        \caption*{\small 0.5 noise – Context}
    \end{minipage}
    \begin{minipage}[b]{0.32\linewidth}
        \centering
        \includegraphics[width=\linewidth, trim=0 0 0 50, clip]{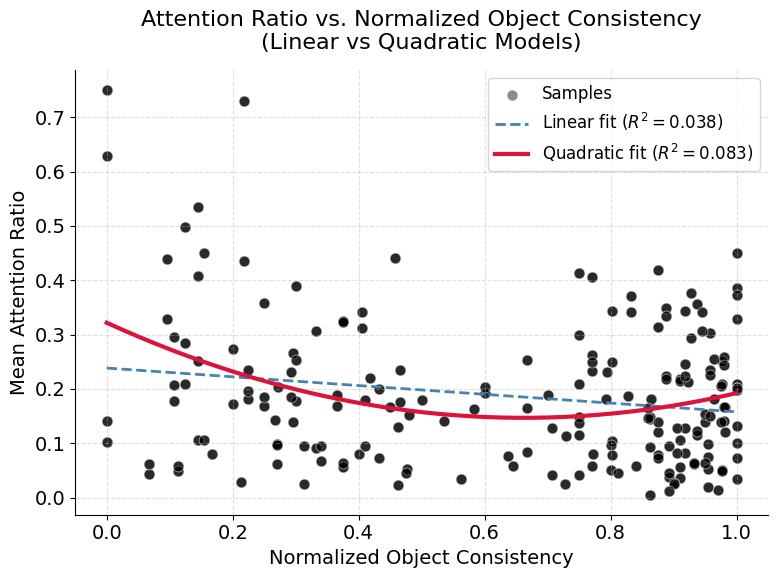}
        \caption*{\small 0.5 noise – All}
    \end{minipage}

    \vspace{10pt}

    \begin{minipage}[b]{0.32\linewidth}
        \centering
        \includegraphics[width=\linewidth, trim=0 0 0 50, clip]{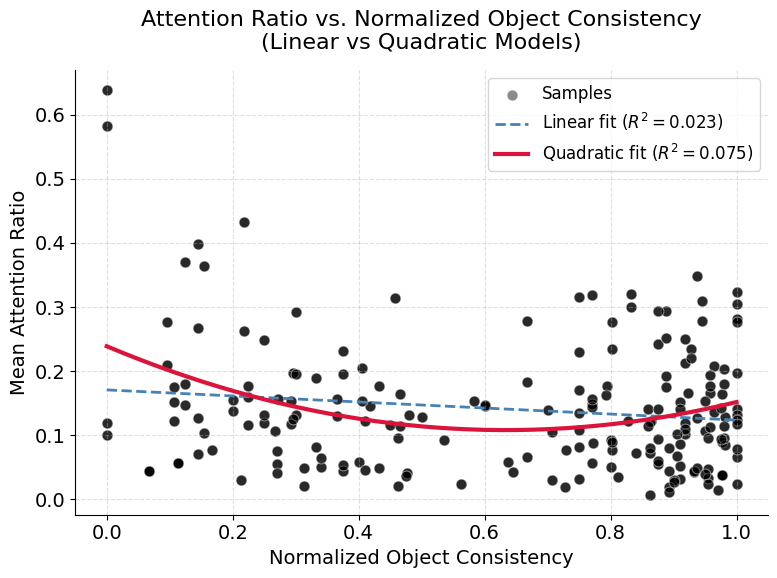}
        \caption*{\small 1 noise – Target}
    \end{minipage}
    \begin{minipage}[b]{0.32\linewidth}
        \centering
        \includegraphics[width=\linewidth, trim=0 0 0 50, clip]{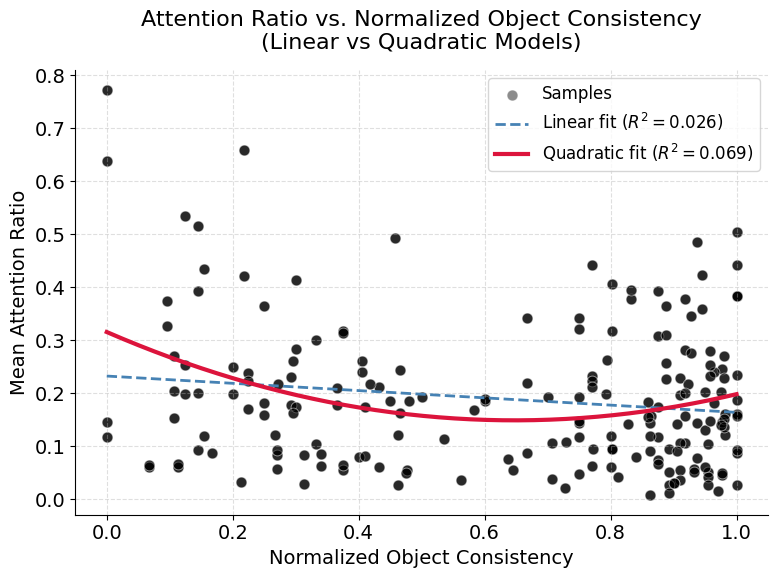}
        \caption*{\small 1 noise – Context}
    \end{minipage}
    \begin{minipage}[b]{0.32\linewidth}
        \centering
        \includegraphics[width=\linewidth, trim=0 0 0 50, clip]{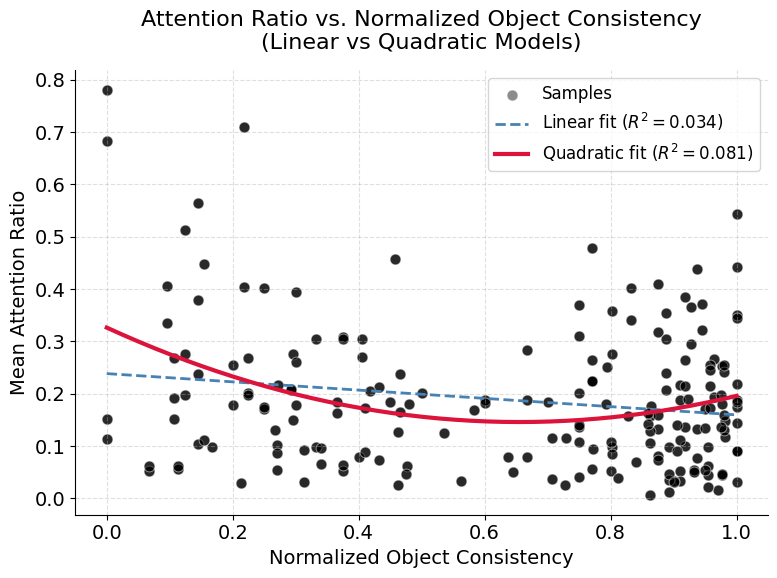}
        \caption*{\small 1 noise – All}
    \end{minipage}

    \caption{\footnotesize Mean attention-to-target ratio as a function of semantic fit for COOCo and VISIONS, at two noise levels (0.5, 1.0) and noise area conditions (target, context, all).}
    \label{fig:attn_continuous_all}
\end{figure*}

\begin{table*}[h]
\footnotesize
\centering
\caption{\footnotesize Accuracy results for LLaVA-OneVision-0.5B on the COOCo dataset across relatedness levels, noise intensities, and noise areas.}
\label{tab:cooco-acc}
\begin{tabular}{l c c c c c}
\hline
Rel. Level & Noise Level & Noise Area & Total Samples & Correct Samples & Accuracy (\%) \\
\hline
low    & 0.0 & --     & 111384 & 41860 & 37.58 \\
     & 0.5 & all     & 111384 & 10816 & 9.71  \\
     &   & context & 111384 & 41548 & 37.30 \\
     &   & target  & 111384 & 4732  & 4.25  \\
     & 1.0 & all     & 111384 & 3302  & 2.96  \\
     &   & context & 111384 & 41990 & 37.70 \\
     &   & target  & 111384 & 728   & 0.65  \\
medium & 0.0 & --     & 117494 & 44954 & 38.26 \\
  & 0.5 & all     & 117494 & 17810 & 15.16 \\
  &   & context & 117494 & 45448 & 38.68 \\
  &   & target  & 117494 & 11674 & 9.94  \\
  & 1.0 & all     & 117494 & 6084  & 5.18  \\
  &   & context & 117494 & 43134 & 36.71 \\
  &   & target  & 117494 & 3510  & 2.99  \\
high   & 0.0 & --     & 34086  & 17654 & 51.79 \\
    & 0.5 & all     & 34086  & 8242  & 24.18 \\
    &   & context & 34086  & 17758 & 52.10 \\
    &   & target  & 34086  & 5252  & 15.41 \\
    & 1.0 & all     & 34086  & 2938  & 8.62  \\
    &   & context & 34086  & 16692 & 48.97 \\
    &   & target  & 34086  & 1820  & 5.34  \\
same target & 0.0 & --     & 40326  & 34476 & 85.49 \\
  & 0.5 & all     & 40326  & 23452 & 58.16 \\
  &   & context & 40326  & 34736 & 86.14 \\
  &   & target  & 40326  & 21008 & 52.10 \\
  & 1.0 & all     & 40326  & 10192 & 25.27 \\
  &   & context & 40326  & 33722 & 83.62 \\
  &   & target  & 40326  & 8970  & 22.24 \\
original & 0.0 & --     & 41626  & 34840 & 83.70 \\
  & 0.5 & all     & 41626  & 19734 & 47.41 \\
  &   & context & 41626  & 37024 & 88.94 \\
  &   & target  & 41626  & 18434 & 44.28 \\
  & 1.0 & all     & 41626  & 8320  & 19.99 \\
  &   & context & 41626  & 35256 & 84.70 \\
  &   & target  & 41626  & 9490  & 22.80 \\
\hline
\end{tabular}
\end{table*}

\begin{table*}[h]
\centering
\footnotesize
\caption{\footnotesize Accuracy results for LLaVA-OneVision-0.5B on the VISIONS dataset across congruency conditions, noise intensities, and noise areas for incongruent and congruent conditions.}
\label{tab:visions-acc}
\begin{tabular}{l c c c c c}
\hline
Rel. Level & Noise Level & Noise Area & Total Samples & Correct Samples & Accuracy (\%) \\
\hline
Incongruent & 0.0 & --     & 8580 & 1950 & 22.73 \\
    & 0.5 & All     & 8580 & 104  & 1.21  \\
    &   & Context & 8580 & 1742 & 20.30 \\
    &   & Target  & 8580 & 52   & 0.61  \\
    & 1.0 & All     & 8580 & 104  & 1.21  \\
    &   & Context & 8580 & 1898 & 22.12 \\
    &   & Target  & 8580 & 52   & 0.61  \\
Congruent   & 0.0 & --     & 8580 & 3250 & 37.88 \\
    & 0.5 & All     & 8580 & 390  & 4.55  \\
    &   & Context & 8580 & 2626 & 30.61 \\
    &   & Target  & 8580 & 624  & 7.27  \\
    & 1.0 & All     & 8580 & 78   & 0.91  \\
    &   & Context & 8580 & 2834 & 33.03 \\
    &   & Target  & 8580 & 624  & 7.27  \\
\hline
\end{tabular}
\end{table*}

\end{document}